\documentclass[a4paper,twoside]{article}

\usepackage{epsfig}
\usepackage{subcaption}
\usepackage{calc}
\usepackage{amssymb}
\usepackage{amstext}
\usepackage{amsmath}
\usepackage{amsthm}
\usepackage{multicol}
\usepackage{pslatex}
\usepackage{apalike}
\usepackage{algorithm2e}
\usepackage[bottom]{footmisc}

\usepackage{graphicx}
\usepackage{booktabs}
\usepackage{multirow}
\usepackage{comment}
\usepackage{colortbl}
\usepackage{amsfonts}
\usepackage{tikz}
\usetikzlibrary{3d,decorations.text,shapes.arrows,positioning,fit,backgrounds,arrows.meta}
\usepackage{transparent}

\usepackage{SCITEPRESS}     

\usepackage[pagebackref=true,breaklinks=true,colorlinks,bookmarks=false]{hyperref}

\newcommand{\e}[1]{\mathrm{e}^{#1}}
\newcommand{\IoU}{\text{IoU}}
\newcommand{\auroc}{\text{AuROC}}
\newcommand{\ause}{\text{AuSE}}
\newcommand{\ece}{\text{ECE}}
\newcommand{\pgnuni}{\text{PGN}_\mathit{uni}}
\newcommand{\pgnoh}{\text{PGN}_\mathit{oh}}
\DeclareMathOperator*{\argmax}{arg\, max}

\definecolor{gainsboro}{rgb}{0.86, 0.86, 0.86}
\definecolor{columbiablue}{rgb}{0.61, 0.87, 1.0}
\definecolor{vividtangerine}{rgb}{1.0, 0.63, 0.54}
\definecolor{grannysmithapple}{rgb}{0.66, 0.89, 0.63}

\begin{document}

\title{Pixel-wise Gradient Uncertainty for Convolutional Neural Networks applied to Out-of-Distribution Segmentation}

\author{\authorname{Kira Maag* and Tobias Riedlinger*}
\affiliation{Technical University of Berlin, Germany}
\email{\{maag, riedlinger\}@tu-berlin.de}
}

\keywords{Deep Learning, Semantic Segmentation, Gradient Uncertainty, Out-of-Distribution Detection.}

\abstract{In recent years, deep neural networks have defined the state-of-the-art in semantic segmentation where their predictions are constrained to a predefined set of semantic classes. They are to be deployed in applications such as automated driving, although their categorically confined expressive power runs contrary to such open world scenarios. Thus, the detection and segmentation of objects from outside their predefined semantic space, i.e., out-of-distribution (OoD) objects, is of highest interest. Since uncertainty estimation methods like softmax entropy or Bayesian models are sensitive to erroneous predictions, these methods are a natural baseline for OoD detection. Here, we present a method for obtaining uncertainty scores from pixel-wise loss gradients which can be computed efficiently during inference. Our approach is simple to implement for a large class of models, does not require any additional training or auxiliary data and can be readily used on pre-trained segmentation models. Our experiments show the ability of our method to identify wrong pixel classifications and to estimate prediction quality at negligible computational overhead. In particular, we observe superior performance in terms of OoD segmentation to comparable baselines on the SegmentMeIfYouCan benchmark, clearly outperforming other methods.}

\onecolumn \maketitle \normalsize \setcounter{footnote}{0} \vfill

\let\thefootnote\relax\footnotetext{* equal contribution}
%
%
%
\section{\uppercase{Introduction}}
\begin{figure}[t]
    \center
    \includegraphics[trim=250 330 300 60,clip,width=0.46\textwidth]{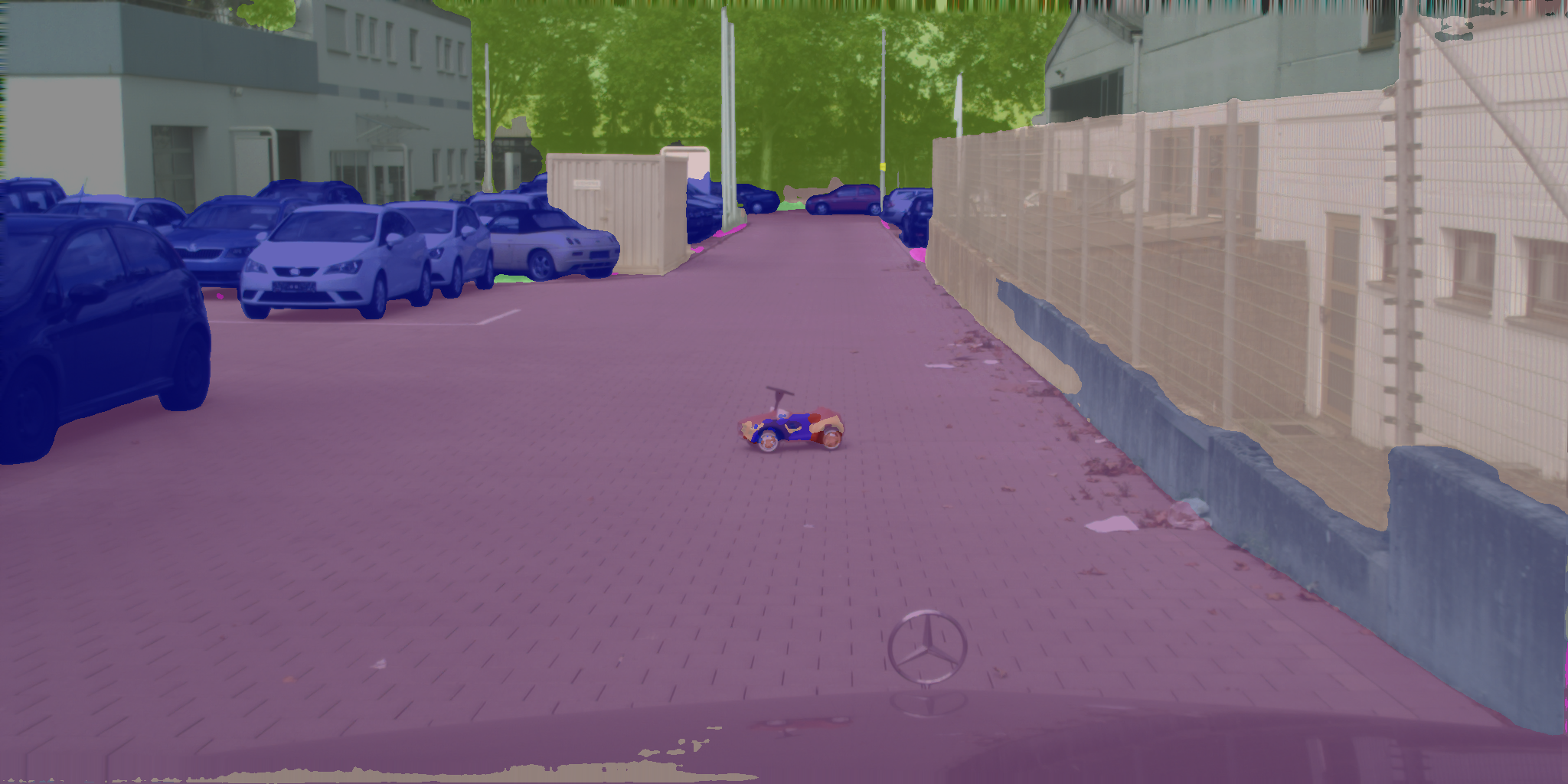}
    \includegraphics[trim=250 330 300 60,clip,width=0.46\textwidth]{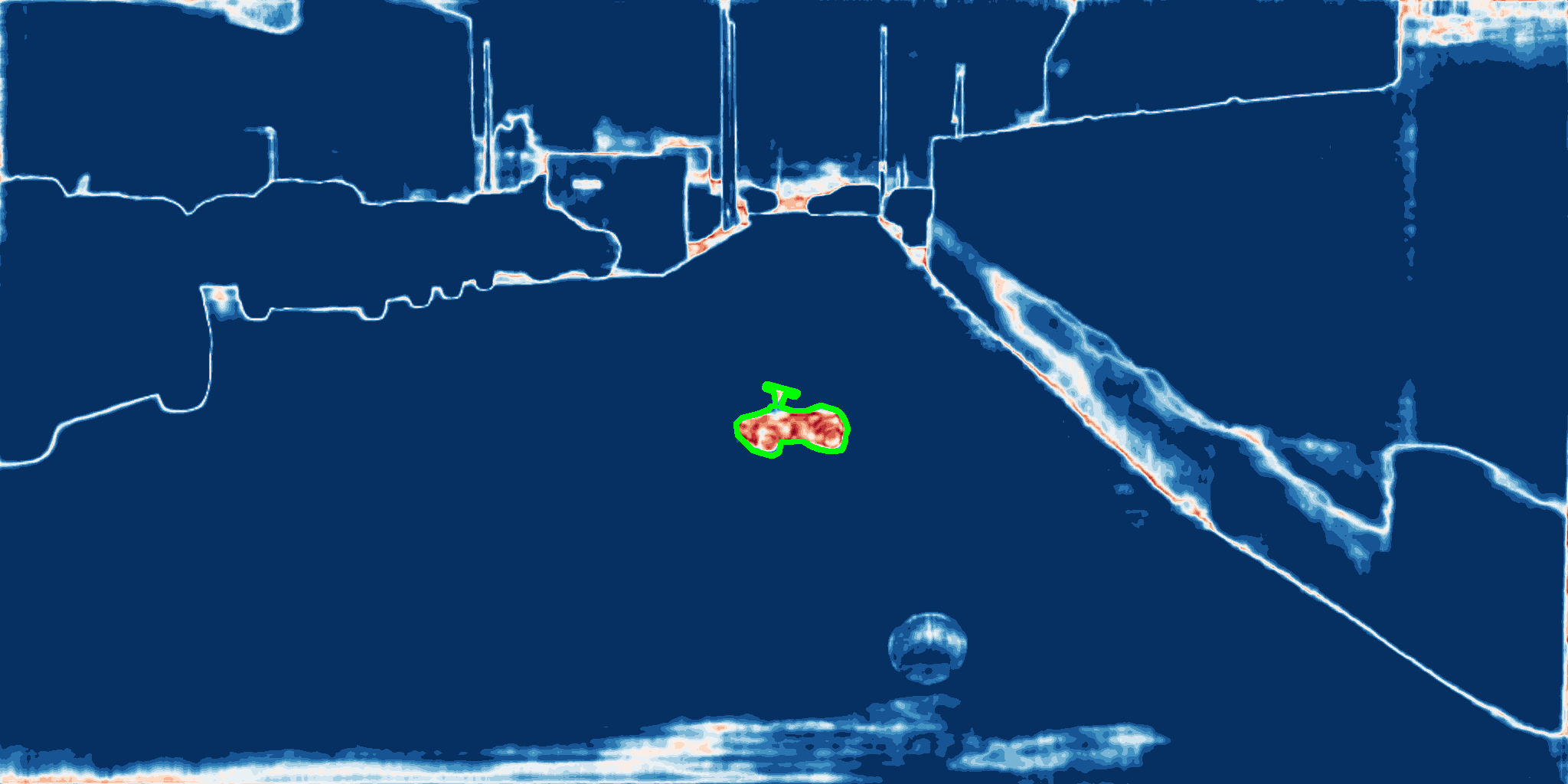}
    \caption{{Top}: Semantic segmentation by a deep neural network. {Bottom}: Gradient uncertainty heatmap obtained by our method.}
    \label{fig:seg_ood}
\end{figure}

Semantic segmentation decomposes the pixels of an input image into segments which are assigned to a fixed and predefined set of semantic classes. 
In recent years, deep neural networks (DNNs) have performed excellently in this task~\cite{Chen2018,Wang2021}, providing comprehensive and precise information about the given scene. 
However, in safety-relevant applications like automated driving where semantic segmentation is used in open world scenarios, DNNs often fail to function properly on unseen objects for which the network has not been trained, see for example the bobby car in Figure~\ref{fig:seg_ood} (top). 
These objects from outside the network's semantic space are called \emph{out-of-distribution} (OoD) objects. 
It is of highest interest that the DNN identifies such objects and abstains from deciding on the semantic class for those pixels covered by the OoD object. 
Another case are OoD objects which might belong to a known class, however, appearing differently to substantial significance from other objects of the same class seen during training. 
Consequently, the respective predictions are prone to error. 
For these objects, as for classical OoD objects, marking them as OoD is preferable to the likely case of misclassification which may happen with high confidence.
This additional classification task should not substantially degrade the semantic segmentation performance itself outside the OoD region. 
The computer vision tasks of identifying and segmenting those objects is called \emph{OoD segmentation}~\cite{Chan2021_1,Maag2022}.

The recent contributions to the emerging field of OoD segmentation are mostly focused on OoD training, i.e., the incorporation of additional training data (not necessarily from the real world), sometimes obtained by large reconstruction models~\cite{Biase2021,Lis2019}. 
Another line of research is the use of uncertainty quantification methods such as Bayesian models~\cite{Mukhoti2020} or maximum softmax probability~\cite{Hendrycks2016}. 
Gradient-based uncertainties are considered for OoD detection in the classification task \cite{Huang2021,Lee2022,Oberdiek2018} but up to now, have not been applied to OoD segmentation. 
In \cite{Grathwohl2020}, it is shown that gradient norms perform well in discriminating between in- and out-of-distribution. 
Moreover, gradient-based features are studied for object detection to estimate the prediction quality in \cite{Riedlinger2023}.
In \cite{Hornauer2022}, loss gradients w.r.t.\ feature activations in monocular depth estimation are investigated and show correlations of gradient magnitude with depth estimation accuracy.

In this work, we introduce a new method for uncertainty quantification in semantic segmentation based on gradient information. 
Magnitude features of gradients can be computed at inference time and provide information about the uncertainty propagated in the corresponding forward pass. 
The features represent pixel-wise uncertainty scores applicable to prediction quality estimation and OoD segmentation. 
An exemplary gradient uncertainty heatmap can be found in Figure~\ref{fig:seg_ood} (bottom). 
Calculating gradient uncertainty scores does not require any re-training of the DNN or computationally expensive sampling. 
Instead, only one backpropagation step for the gradients with respect to the final convolutional network layer is performed per inference to produce gradient scores. 
Note, that more than one backpropagation step can be performed to compute deeper gradients which considers other parameters of the model architecture. 
An overview of our approach is shown in Figure~\ref{fig: method illustration}. 

An application to dense predictions such as semantic segmentation has escaped previous research, potentially due to the superficial computational overhead.
Single backpropagation steps per pixel on high-resolution input images quickly become infeasible given that around $10^6$ gradients have to be calculated. 
To overcome this issue, we present a new approach to exactly compute the pixel-wise gradient scores in a batched and parallel manner applicable to a large class of segmentation architectures.
This is possible due to a convenient factorization of \(p\)-norms for appropriately factorizing tensors, such as the gradients for convolutional neural networks.
We use the computed gradient scores to estimate the model uncertainty on pixel-level and also the prediction quality on segment-level. 
Moreover, the gradient uncertainty heatmaps are investigated for OoD segmentation where high scores indicate possible OoD objects. 
We demonstrate the efficiency of our method in explicit runtime measurements and show that the computational overhead introduced is marginal compared with the forward pass.
This suggests that our method is applicable in extremely runtime-restricted applications of semantic segmentation.

For our method, we only assume a pre-trained semantic segmentation network ending on a convolutional layer which is a common case. 
In our experiments, we employ a state-of-the-art semantic segmentation network~\cite{Chen2018} trained on Cityscapes~\cite{Cordts2016} evaluating in-distribution uncertainty estimation.
We demonstrate OoD detection performance on four well-known OoD segmentation datasets, namely LostAndFound~\cite{Pinggera2016}, Fishyscapes~\cite{Blum2019}, RoadAnomaly21 and RoadObstacle21~\cite{Chan2021_1}.
The source code of our method is made publicly available at \url{https://github.com/tobiasriedlinger/uncertainty-gradients-seg}. 
We summarize our contributions as follows:
\begin{itemize}
    \item We introduce a new gradient-based method for uncertainty quantification in semantic segmentation. 
    This approach is applicable to a wide range of common segmentation architectures.
    \item For the first time, we show an efficient way of computing gradient norms in semantic segmentation on the pixel-level in a parallel manner making our method far more efficient than sampling-based methods which is demonstrated in explicit time measurements.
    \item We demonstrate the effectiveness of our method to predictive error detection and OoD segmentation. 
    For OoD segmentation, we achieve area under precision-recall curve values of up to $69.3\%$ on the LostAndFound benchmark outperforming a variety of methods. 
\end{itemize}
%
%
%
\section{\uppercase{Related work}}\label{sec:rel_work}
%
%
\paragraph{Uncertainty Quantification.} 
Bayesian approaches~\cite{Mackay1992} are widely used to estimate model uncertainty. 
The well-known approximation, MC Dropout~\cite{Gal2016}, has proven to be computationally feasible for computer vision tasks and has also been applied to semantic segmentation~\cite{Lee2020}. 
In addition, this method is considered to filter out predictions with low reliability~\cite{Wickstrom2018}. 
In \cite{Blum2019}, pixel-wise uncertainty estimation methods are benchmarked based on Bayesian models or the network's softmax output. 
Uncertainty information is extracted on pixel-level by using the maximum softmax probability and MC Dropout in \cite{Hoebel2020}.
Prediction quality evaluation approaches were introduced in \cite{DeVries2018,Huang2016} and work on single objects per image.
These methods are based on additional CNNs acting as post-processing mechanism.
The concepts of meta classification (false positive / FP detection) and meta regression (performance estimation) on segment-level 
were introduced in \cite{Rottmann2018}. 
This line of research has been extended by a temporal component~\cite{Maag2019} and transferred to object detection~\cite{Schubert2020,Riedlinger2023} as well as to instance segmentation~\cite{Maag2020,Maag2021}.

While MC Dropout as a sampling approach is still computationally expensive to create pixel-wise uncertainties, our method computes only the gradients of the last layer during a single inference run and can be applied to a wide range of semantic segmentation networks without architectural changes.
Compared with the work presented in \cite{Rottmann2018}, our gradient information can extend the features extracted from the segmentation network's softmax output to enhance the segment-wise quality estimation.
%
%

\paragraph{OoD Segmentation.} 
Uncertainty quantification methods demonstrate high uncertainty for erroneous predictions, so they are often applied to OoD detection. 
For instance, this can be accomplished via maximum softmax (probability)~\cite{Hendrycks2016}, MC Dropout~\cite{Mukhoti2020} or deep ensembles~\cite{Lakshminarayanan2017} the latter of which also capture model uncertainty by averaging predictions over multiple sets of parameters in a Bayesian manner. 
Another line of research is OoD detection training, relying on the exploitation of additional training data, not necessarily from the real world, but disjoint from the original training data~\cite{Blum2019_1,Chan2021,Grcic2022,Grcic2023,Liu2023,Nayal2023,Rai2023,Tian2022}. 
In this regard, an external reconstruction model followed by a discrepancy network is considered in \cite{Biase2021,Lis2019,Lis2020,Vojir2021,Vojir2023} and normalizing flows are leveraged in \cite{Blum2019_1,Grcic2021,Gudovskiy2023}. 
In \cite{Lee2018,Liang2018}, adversarial perturbations are performed on the input images to improve the separation of in- and out-of-distribution samples.

Specialized training approaches for OoD detection are based on different kinds of re-training with additional data and often require generative models.
Meanwhile, our method does not require OoD data, re-training or complex auxiliary models. 
Moreover, we do not run a full backward pass which is, however, required for the computation of adversarial samples.
In fact we found that it is often sufficient to only compute the gradients of the last convolutional layer.
Our method is more related to classical uncertainty quantification approaches like maximum softmax, MC Dropout and ensembles. 
Note that the latter two are based on sampling and thus, much more computationally expensive compared to single inference. 
%
%
%
\section{\uppercase{Method description}}\label{sec:method}
In the following, we consider a neural network with parameters $\theta$ yielding classification probabilities $\hat{\pi}(x, \theta) = (\hat{\pi}_1, \ldots, \hat{\pi}_C)$ 
over $C$ semantic categories when presented with an input $x$.
During training on paired data $(x, y)$, where $y \in \{1, \ldots, C\}$ is the semantic label given to $x$, such a model is commonly trained by minimizing some kind of loss function $\mathcal{L}$ between $y$ and the predicted probability distribution $\hat{\pi}(x, \theta)$ using gradient descent on $\theta$.
The gradient step $\nabla_{\! \theta} \mathcal{L}(\hat{\pi}(x, \theta) \| y)$ then indicates the direction and strength of training feedback obtained by $(x, y)$.
Asymptotically (in the amount of data and training steps taken), the expectation \(
    \mathbb{E}_{(X, Y) \sim P} [\nabla_{\!\theta} \mathcal{L}(\hat{\pi}(X, \theta) \| Y)]
\)
of the gradient w.r.t.\ the data generating distribution $P$ will vanish since $\theta$ sits at a local minimum of $\mathcal{L}$.
We assume, that strong models which achieve high test accuracy can be seen as an approximation of such a parameter configuration $\theta$.
Such a model has small gradients on in-distribution data which is close to samples $(x, y)$ in the training data.
Samples that differ from training data may receive larger feedback.
Like-wise, it is plausible to obtain large training gradients from OoD samples that are far away from the effective support of $P$.

In order to quantify uncertainty about the prediction $\hat{\pi}(x, \theta)$, we replace the label $y$ from above by some fixed auxiliary label for which we make two concrete choices in our method.
On the one hand, we replace $y$ by the class prediction one-hot vector $y^\mathit{oh}_k = \delta_{k \hat{c}}$ with $\hat{c} = \mathrm{arg\,max}_{k = 1, \ldots, C} \, \hat{\pi}_k$ and the Kronecker symbol $\delta_{ij}$.
This quantity correlates strongly with training labels $y$ on in-distribution data.
On the other hand, we regard a constant uniform, all-warm label $y^\mathit{uni}_k = 1/C$ for $k = 1, \ldots, C$ as a replacement for $y$.
To motivate the latter choice, we consider that classification models are oftentimes trained on the categorical cross entropy loss 
\begin{equation}
    \label{eq: cross entropy}
    \mathcal{L}(\hat{\pi} \| y) = - \sum_{k = 1}^C y_k \log(\hat{\pi}_k).
\end{equation}
Since the gradient of this loss function is linear in the label $y$, a uniform choice \(y^\mathit{uni}\) will return the average gradient per class which is expected to be large on OoD data where all possible labels are similarly unlikely.
The magnitude of $\nabla_{\! \theta}\mathcal{L}(\hat{\pi} \| y)$ serves as uncertainty / OoD score.
In the following, we explain how to compute such scores for pixel-wise classification models.

%
%
\subsection{Efficient Computation in Semantic Segmentation}\label{sec:method_grads}
\begin{figure*}
    \centering
    \resizebox{0.7\linewidth}{!}{\includegraphics{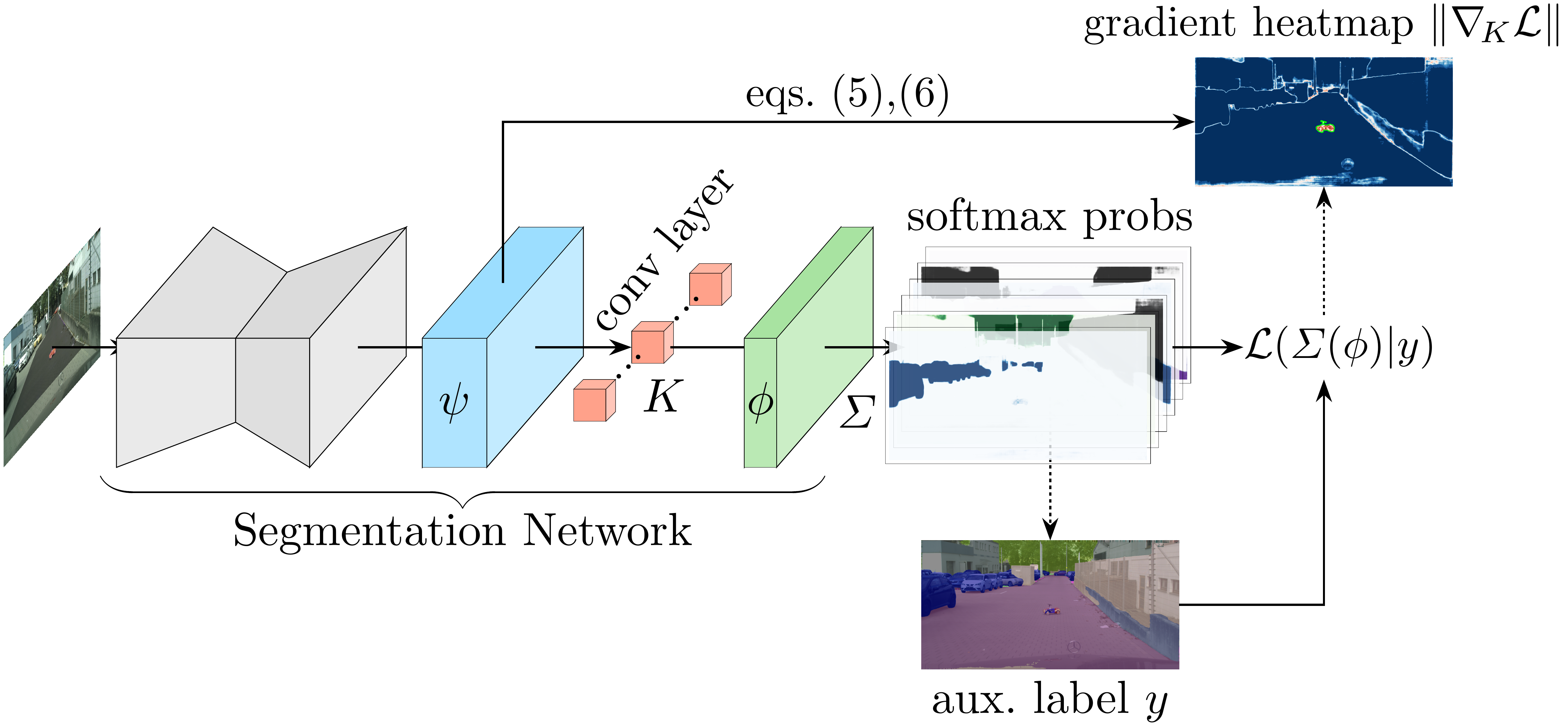}}
    \caption{Schematic illustration of the computation of pixel-wise gradient norms for a semantic segmentation network with a final convolution layer.
    Auxiliary labels may be derived from the softmax prediction or supplied in any other way (e.g., as a uniform all-warm label).
    We circumvent direct back propagation per pixel by utilizing eqs.~\eqref{eq: explicit gradient computation},~\eqref{eq: uniform gradients}.}
    \label{fig: method illustration}
\end{figure*}
We regard a generic segmentation architecture utilizing a final convolution as the pixel-wise classification mechanism.
In the following, we also assume that the segmentation model is trained via the commonly-used pixel-wise cross entropy loss $\mathcal{L}_{ab}(\phi(\theta) | y)$ (cf.\ eq.~\eqref{eq: cross entropy}) over each pixel $(a,b)$ with feature map activations $\phi(\theta)$.
However, similarly compact formulas hold for arbitrary loss functions.
Pixel-wise probabilities are obtained by applying the softmax function $\varSigma$ to each output pixel, i.e., $\hat{\pi}_{ab, k} = \varSigma^k\left(\phi_{ab}(\theta)\right)$. 
With eq.~\eqref{eq: cross entropy}, we find for the loss gradient 
\begin{equation}
    \label{eq: cross entropy gradient}
    \nabla_{\!\theta} \mathcal{L}_{ab}(\varSigma(\phi(\theta)) \| y) 
    = \sum_{k = 1}^C \varSigma^k(\phi_{ab}) (1 - y_k) \cdot \nabla_{\!\theta} \phi^k_{ab}(\theta).
\end{equation}
Here, $\theta$ is any set of parameters within the neural network.
Detailed derivations of this and the following formulas of this section are provided in Appendix~\ref{sec:app_A}.
Here, $\phi$ is the convolution result of a pre-convolution feature map $\psi$ (see Figure~\ref{fig: method illustration}) against a filter bank $K$ which assumes the role of $\theta$.
$K$ has parameters $(K_e^h)_{fg}$ where $e$ and $h$ indicate in- and out-going channels respectively and $f$ and $g$ index the spatial filter position.
The features $\phi$ are linear in both, $K$ and $\psi$, and explicitly depend on bias parameters $\beta$ in the form
\begin{equation}
    \phi_{ab}^d
    = 
    (K \ast \psi)_{ab}^d + \beta^d
    = 
    \sum_{j = 1}^{\kappa} \sum_{p,q=-s}^s (K_j^d)_{pq} \psi^j_{a + p, b + q} + \beta^d.
\end{equation}
We denote by $\kappa$ the total number of in-going channels and $s$ is the spatial extent to either side of the filter $K_j^d$ which has total size $(2s+1) \times (2s+1)$.
We make different indices in the convolution explicit in order to compute the exact gradients w.r.t.\ the filter weights \((K_e^h)_{fg}\), where we find for the last layer gradients
\begin{equation}
    \frac{\partial \phi_{ab}^d}{\partial (K_e^h)_{fg}} = \delta_{dh} \psi_{a+f, b+g}^e.
\end{equation}
Together with eq.~\eqref{eq: cross entropy gradient}, we obtain the closed form for computing the correct backpropagation gradients of the loss on pixel-level for our choices of auxiliary labels which we state in the following paragraph.
The computation of the loss gradient can be traced in Figure~\ref{fig: method illustration}.

If we take the predicted class $\hat{c}$ as a one-hot label per pixel as auxiliary labels, i.e., $y^\mathit{oh}$, we obtain for the last layer gradients
\begin{equation}
    \label{eq: explicit gradient computation}
    \frac{\partial \mathcal{L}_{ab}(\varSigma(\phi) \| y^\mathit{oh})}{\partial K_e^h} = \varSigma^h(\phi_{ab}) \cdot (1 - \delta_{h \hat{c}}) \cdot \psi_{ab}^e
\end{equation}
which depends on quantities that are easily accessible during a forward pass through the network.
Note, that the filter bank $K$ for the special case of ($1 \times 1$)-convolutions does not require spatial indices which is a common situation in segmentation networks, albeit, not necessary for our method to be applied.
Similarly, we find for the uniform label $y^\mathit{uni}_j = 1 / C$
\begin{equation}
    \label{eq: uniform gradients}
    \frac{\partial \mathcal{L}_{ab}(\varSigma(\phi) \| y^{\mathit{uni}})}{\partial K_e^h}
    =
    \frac{C - 1}{C} \varSigma^h(\phi_{ab}) \psi_{ab}^e.
\end{equation}
These formulas reveal a practical factorization of the gradient which we will exploit computationally in the next section.
Therefore, pixel-wise gradient norms are simple to implement and particularly efficient to compute for the last layer of the segmentation model. 
In Appendix~\ref{sec:deep_grad}, we cover formulas for the more general case of deeper gradients, as well. 
%
%
\subsection{Uncertainty Scores}\label{sec:method_ood}
We obtain pixel-wise scores, i.e., still depending on $a$ and $b$, by computing the partial norm $\|\nabla_{\! K} \mathcal{L}_{ab}\|_p$ of this tensor over the indices $e$ and $h$ for some fixed value of $p$.
This can again be done in a memory efficient way by the natural decomposition $\partial \mathcal{L} / \partial K_e^h = S^h \cdot \psi^e$.
In addition to their use in error detection, these scores can be used in order to detect OoD objects in the input, i.e., instances of categories not present in the training distribution of the segmentation network.
We identify OoD regions with pixels that have a gradient score higher than some threshold and find connected components like the one shown in Figure~\ref{fig:seg_ood} (bottom).
Different values of $p$ are studied in the appendix.
We also consider values $0 < p < 1$ in addition to positive integer values.
Note, that such choices do not strictly define the geometry of a vector space norm, however, $\|\cdot\|_p$ may still serve as a notion of magnitude and generates a partial ordering.
The tensorized multiplications required in eqs.~\eqref{eq: explicit gradient computation} and \eqref{eq: uniform gradients} are far less computationally expensive than a forward pass through the DNN.
This means that computationally, this method is preferable over prediction sampling like MC Dropout or ensembles.
We abbreviate our method using the \textbf{p}ixel-wise \textbf{g}radient \textbf{n}orms obtained from eqs.~\eqref{eq: explicit gradient computation} and \eqref{eq: uniform gradients} by $\pgnoh$ and $\pgnuni$, respectively.
The particular value of $p$ is denoted by superscript, e.g., $\pgnoh^{p=0.3}$ for the ($p=0.3$)-seminorm of gradients obtained from $y^\mathit{oh}$. 

%
%
%
\section{\uppercase{Experiments}}\label{sec:exp}
In this section, we present the experimental setting first and then evaluate the uncertainty estimation quality of our method on pixel and segment level. 
We apply our gradient-based method to OoD segmentation, show some visual results and explicitly measure the runtime of our method.
%
%
\subsection{Experimental Setting}\label{sec:exp_setting}
%
%
\paragraph{Datasets.} 
We perform our tests on the Cityscapes~\cite{Cordts2016} dataset for semantic segmentation in street scenes and on four OoD segmentation datasets$^1$\let\thefootnote\relax\footnotetext{$^1$Benchmark: \url{http://segmentmeifyoucan.com/}}.
The Cityscapes dataset consists of $2,\!975$ training and $500$ validation images of dense urban traffic in $18$ and $3$ different German towns, respectively.
The LostAndFound (LAF) dataset \cite{Pinggera2016} contains $1,\!203$ validation images with annotations for the road surface and the OoD objects, i.e., small obstacles on German roads in front of the ego-car. A filtered version (LAF test-NoKnown) is provided in \cite{Chan2021_1}. 
The Fishyscapes LAF dataset \cite{Blum2019} includes $100$ validation images (and $275$ non-public test images) and refines the pixel-wise annotations of the LAF dataset distinguishing between OoD object, background (Cityscapes classes) and void (anything else). 
The RoadObstacle21 dataset \cite{Chan2021_1} ($412$ test images) is comparable to the LAF dataset as all obstacles appear on the road, but it contains more diversity in the OoD objects as well as in the situations. 
In the RoadAnomaly21 dataset \cite{Chan2021_1} ($100$ test images), a variety of unique objects (anomalies) appear anywhere on the image which makes it comparable to the Fishyscapes LAF dataset. 
%
%
\paragraph{Segmentation Networks.} 
We consider a state-of-the-art DeepLabv3+ network~\cite{Chen2018} with two different backbones, WideResNet38~\cite{Wu2016} and SEResNeXt50~\cite{Hu2018}. 
The network with each respective backbone is trained on Cityscapes achieving a mean $\IoU$ value of $90.58\%$ for the WideResNet38 backbone and $80.76\%$ for the SEResNeXt50 on the Cityscapes validation set. 
We use one and the same model trained exclusively on the Cityscapes dataset for both tasks, the uncertainty estimation and the OoD segmentation, as our method does not require additional training.
Therefore, our method leaves the entire segmentation performance of the base model completely intact.
%
%
\subsection{Numerical Results}\label{sec:exp_results}
We provide results for both, error detection and OoD segmentation in the following and refer to Appendix~\ref{sec:app_D} for an ablation study on different values for $p$ and to Appendix~\ref{sec:app_E} for further results where the gradients of deeper layers are computed. 

\paragraph{Pixel-wise Uncertainty Evaluation.} 
\begin{table}[t]
\caption{Pixel-wise uncertainty evaluation results for both backbone architectures and the Cityscapes dataset in terms of $\ece$ and $\ause$.}
\centering
\scalebox{0.75}{
\begin{tabular}{l cc cc}
\toprule
& \multicolumn{2}{c}{WideResNet} & \multicolumn{2}{c}{SEResNeXt} \\
\cmidrule(r){2-3} \cmidrule(r){4-5}
 & $\ece$ $\downarrow$ & $\ause$ $\downarrow$ & $\ece$ $\downarrow$ & $\ause$ $\downarrow$ \\
\midrule
Ensemble    & 0.0173 & 0.4543 & 0.0279 & 0.0482 \\
MC Dropout  & 0.0444 & 0.7056 & 0.0091 & 0.5867 \\
Maximum Softmax    & $\mathbf{0.0017}$ & $\underline{0.0277}$ & $\mathbf{0.0032}$ & $\mathbf{0.0327}$ \rule{0mm}{3.5mm}\\
Entropy            & $0.0063$ & $0.0642$ & $0.0066$ & $0.0623$ \rule{0mm}{3.5mm}\\
$\pgnoh^{p=2}$ (ours) & $\underline{0.0019}$ & $\mathbf{0.0268}$ & $\underline{0.0039}$ & $\underline{0.0365}$\rule{0mm}{3.5mm} \\
\bottomrule
\end{tabular} }
\label{tab:eval_pix}
\end{table}
In order to assess the correlation of uncertainty and prediction errors on the pixel level, we resort to the commonly used sparsification graphs~\cite{Ilg2018}.
Sparsification graphs normalized w.r.t.\ the optimal oracle (sparsification error) can be compared in terms of the so-called area under the sparsification error curve ($\ause$).
The closer the uncertainty estimation is to the oracle in terms of Brier score evaluation, i.e., the smaller the $\ause$, the better the uncertainty eliminates false predictions by the model.
The $\ause$ metric is capable of grading the uncertainty ranking, however, does not address the statistics in terms of given confidence.
Therefore, we resort to an evaluation of calibration in terms of expected calibration error ($\ece$,~\cite{Guo2017}) to assess the statistical reliability of the uncertainty measure.

As baseline methods we consider the typically used uncertainty estimation measures, i.e., mutual information computed via samples from deep ensembles and MC dropout as well as the uncertainty ranking provided by the maximum softmax probabilities native to the segmentation model and the softmax entropy. 
An evaluation of calibration errors requires normalized scores, so we normalize our gradient scores according to the highest value obtained on the test data for the computation of $\ece$.

The resulting metrics are compiled in Table~\ref{tab:eval_pix} for both architectures evaluated on the Cityscapes val split. 
We see that the calibration of our method is roughly on par with the stronger maximum softmax baseline (which was trained to exactly that aim via the negative log-likelihood loss) and outperforms the other three baselines. 
In particular, we achieve superior values to MC Dropout which represents the typical uncertainty measure for semantic segmentation. 
%
%
\paragraph{Segment-wise Prediction Quality Estimation.} 
\begin{figure}[t]
    \center
    \includegraphics[width=0.44\textwidth]{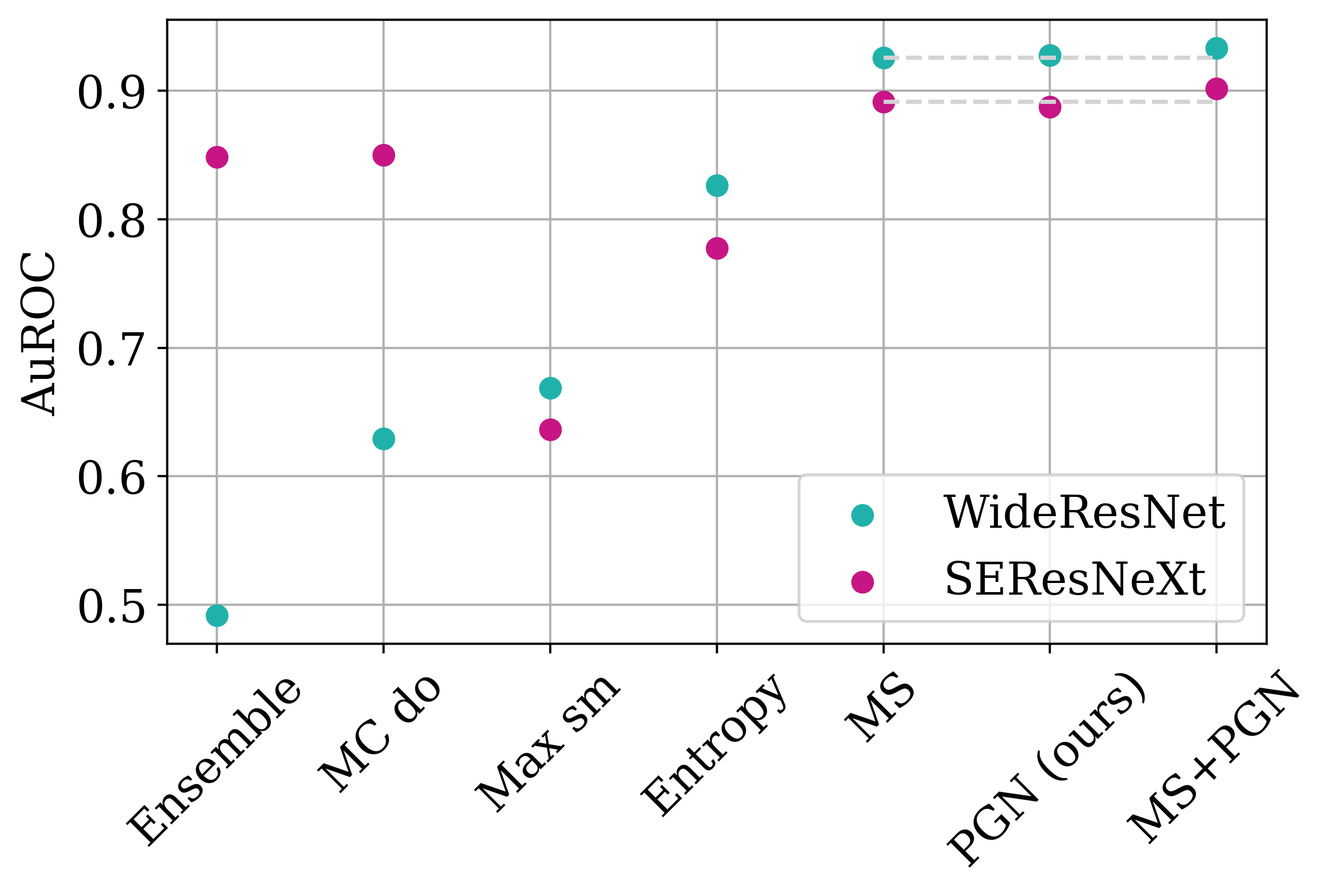}
    \includegraphics[width=0.44\textwidth]{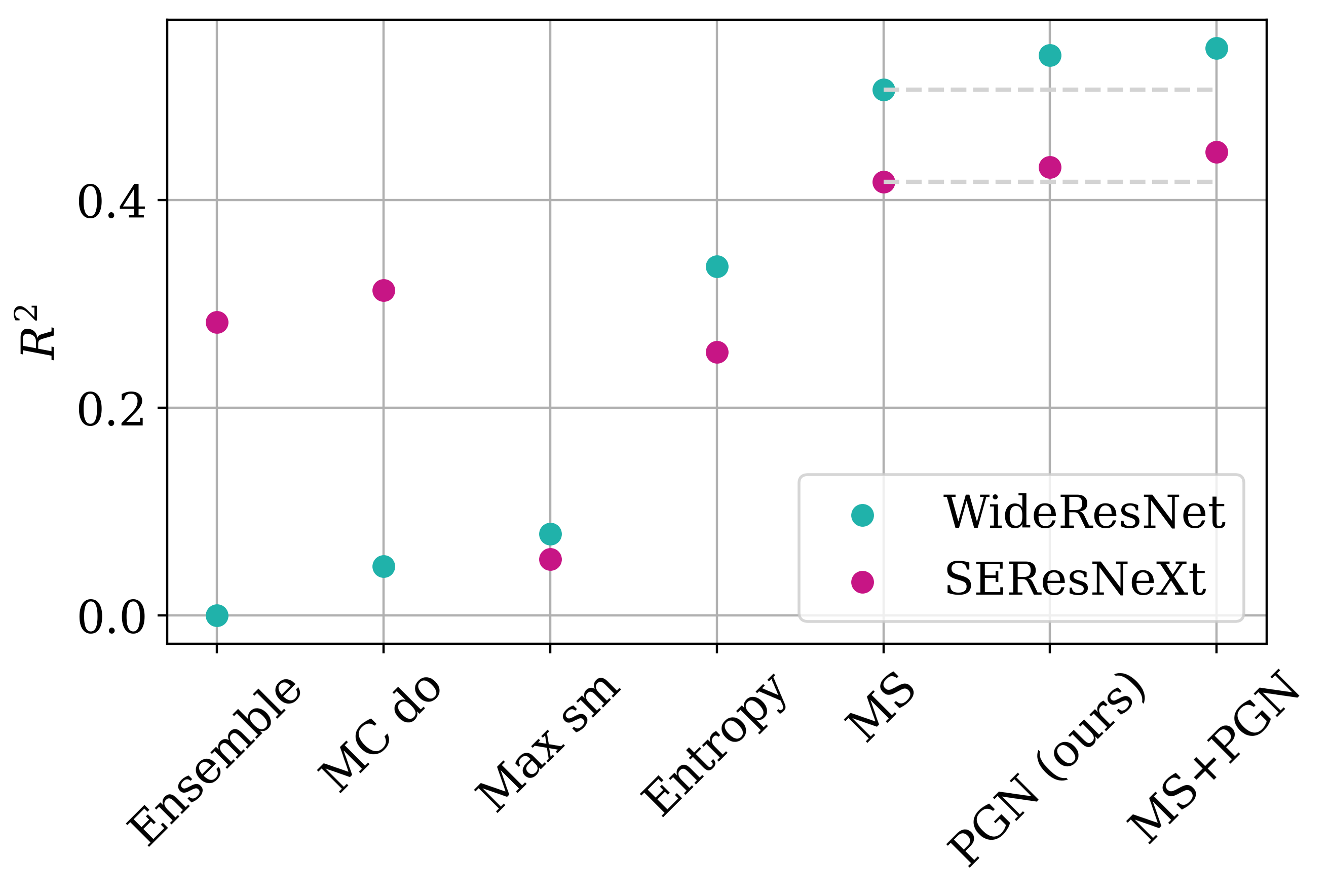}
    \caption{Segment-wise uncertainty evaluation results for both backbone architectures and the Cityscapes dataset in terms of classification $\auroc$ and regression $R^2$ values. From left to right: ensemble, MC Dropout, maximum softmax, mean entropy, MetaSeg (MS) approach, gradient features obtained by predictive one-hot and uniform labels (PGN), MetaSeg in combination with PGN.}
    \label{fig:metaseg}
\end{figure}
\begin{figure*}[t]
    \center
    \includegraphics[trim=309 220 309 0,clip,width=0.24\textwidth]{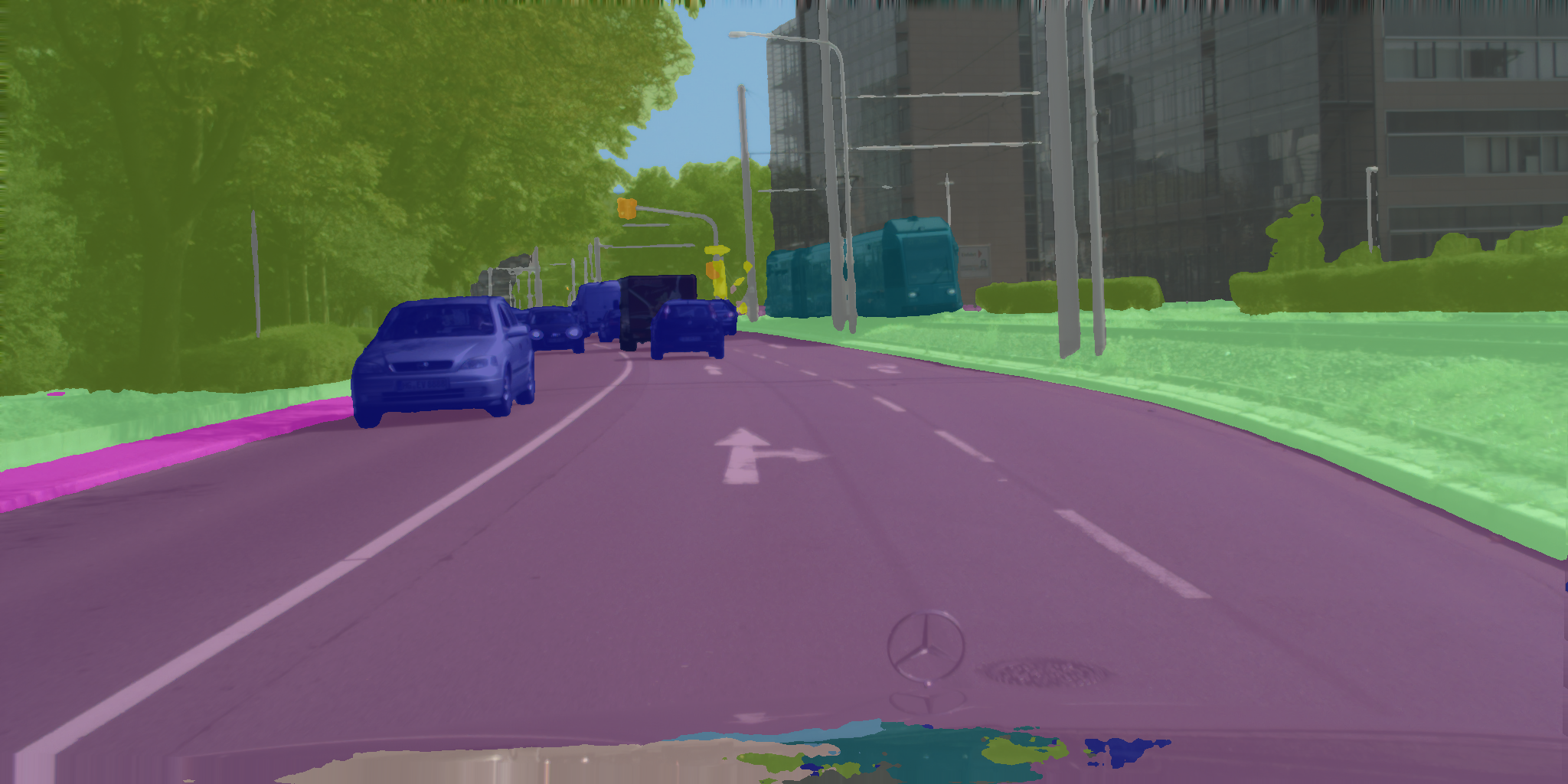}
    \includegraphics[trim=225 160 225 0,clip,width=0.24\textwidth]{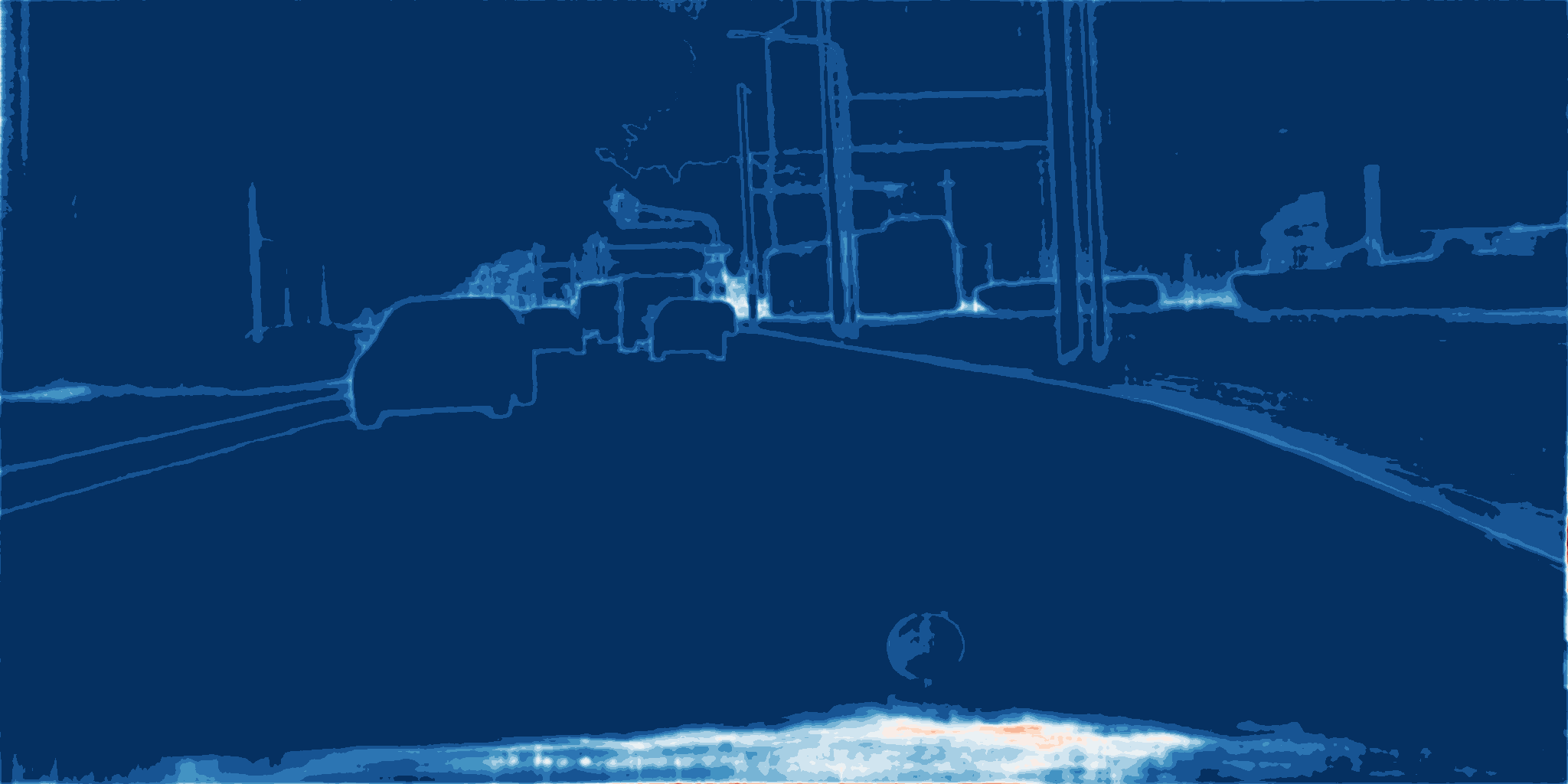}
    \hspace{1ex}
    \includegraphics[width=0.24\textwidth]{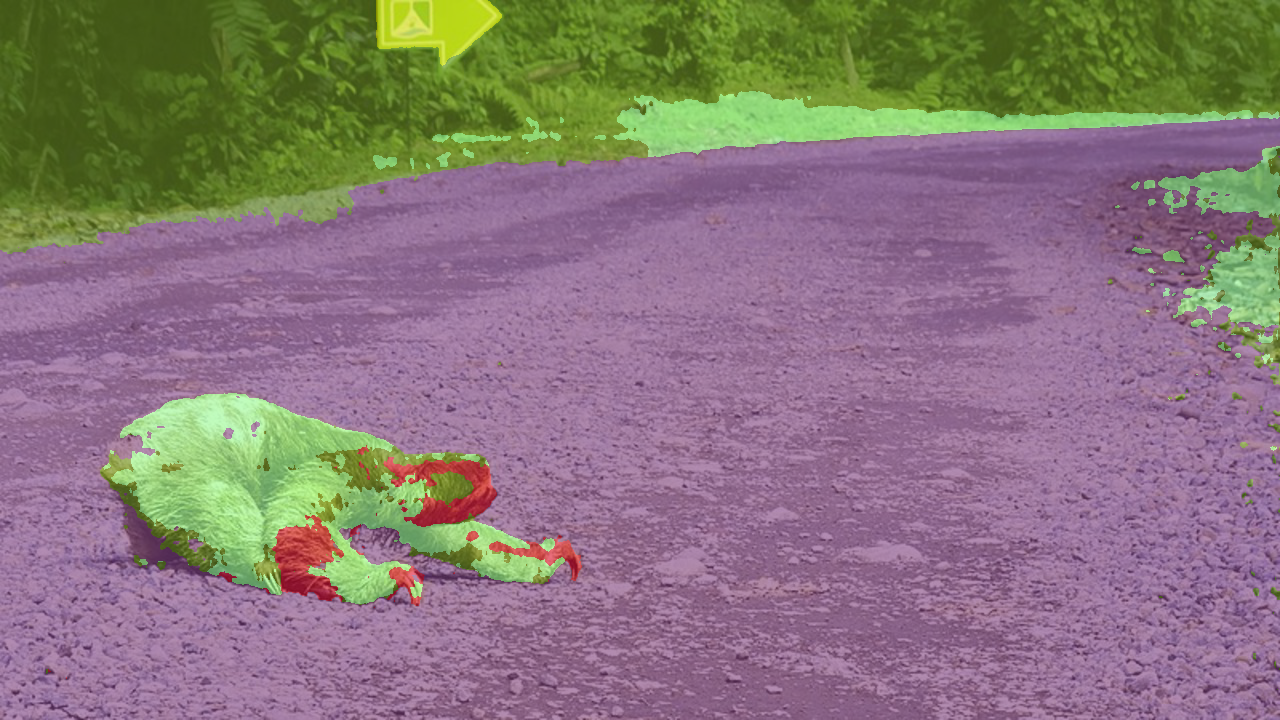}
    \includegraphics[width=0.24\textwidth]{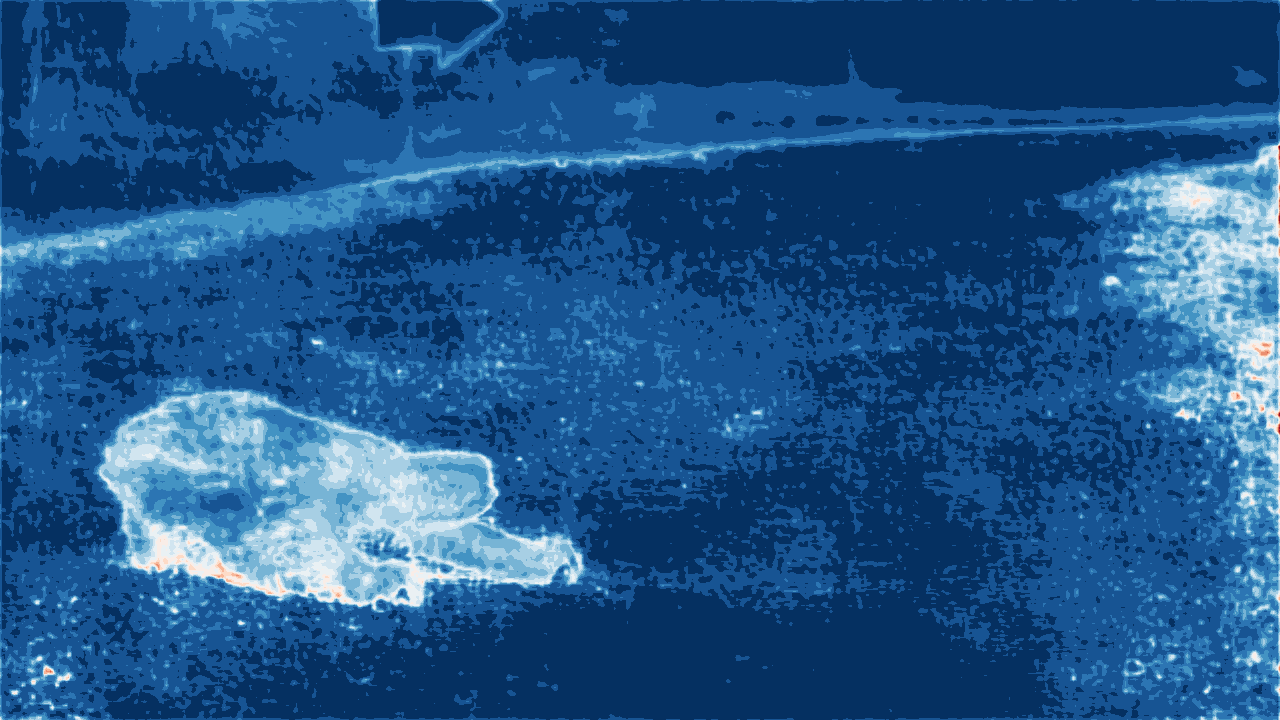}
    \caption{Semantic segmentation prediction and $\mathrm{PGN}_{\mathit{uni}}^{p=0.5}$ heatmap for the Cityscapes dataset ({left}) and the RoadAnomaly21 dataset ({right}) for the WideResNet backbone.}
    \label{fig:heatmaps}
\end{figure*}
\begin{table*}[t]
\caption{OoD segmentation benchmark results for the LostAndFound and the RoadObstacle21 dataset.}
\centering
\resizebox{0.98\linewidth}{!}{
\begin{tabular}{c ccc cc ccc cc ccc}
\toprule 
\multicolumn{1}{c}{} & \multirow{2}{*}{\rotatebox{90}{\footnotesize OoD data}} & \multirow{2}{*}{\rotatebox{90}{\footnotesize OoD arch}} & \multirow{2}{*}{\rotatebox{90}{\footnotesize adversarial}} & \multicolumn{5}{c}{LostAndFound test-NoKnown} & \multicolumn{5}{c}{RoadObstacle21} \\[7pt]
\cmidrule(r){5-9} \cmidrule(r){10-14}
& & & & AuPRC $\uparrow$ & FPR$_{95}$ $\downarrow$ & $\overline{\text{sIoU}}$ $\uparrow$ & $\overline{\text{PPV}}$ $\uparrow$ & $\overline{F_1}$ $\uparrow$ & AuPRC $\uparrow$ & FPR$_{95}$ $\downarrow$ & $\overline{\text{sIoU}}$ $\uparrow$ & $\overline{\text{PPV}}$ $\uparrow$ & $\overline{F_1}$ $\uparrow$\\[3pt]
\midrule
Void Classifier    & $\checkmark$ & & & $4.8$ & $47.0$ & $1.8$ & $35.1$ & $1.9$      & $10.4$ & $41.5$ & $6.3$ & $20.3$ & $5.4$ \\
Maximized Entropy  & $\checkmark$ & ($\checkmark$) & & $77.9$ & $9.7$ & $\mathbf{45.9}$ & $63.1$ & $\mathbf{49.9}$    & $\mathbf{85.1}$ & $\mathbf{0.8}$ & $\mathbf{47.9}$ & $\mathbf{62.6}$ & $\mathbf{48.5}$ \\
SynBoost           & $\checkmark$ & $\checkmark$ & & $\mathbf{81.7}$ & $\mathbf{4.6}$ & $36.8$ & $\mathbf{72.3}$ & $48.7$    & $71.3$ & $3.2$ & $44.3$ & $41.8$ & $37.6$ \\
Image Resynthesis  & & $\checkmark$ & & $57.1$ & $8.8$ & $27.2$ & $30.7$ & $19.2$    & $37.7$ & $4.7$ & $16.6$ & $20.5$ & $8.4$ \\
Embedding Density  & & $\checkmark$ & & $61.7$ & $10.4$ & $37.8$ & $35.2$ & $27.6$   & $0.8$ & $46.4$ & $35.6$ & $2.9$ & $2.3$ \\
ODIN               & & & $\checkmark$ & $52.9$ & $30.0$ & $39.8$ & $49.3$ & $34.5$  & $22.1$ & $15.3$ & $21.6$ & $18.5$ & $9.4$ \\
Mahalanobis        & & & $\checkmark$ & $55.0$ & $12.9$ & $33.8$ & $31.7$ & $22.1$  & $20.9$ & $13.1$ & $13.5$ & $21.8$ & $4.7$ \\
\midrule 
Ensemble           & & ($\checkmark$) & & $2.9$ & $82.0$ & $6.7$ & $7.6$ & $2.7$      & $1.1$ & $77.2$ & $8.6$ & $4.7$ & $1.3$ \\
MC Dropout         & & ($\checkmark$) & & $36.8$ & $35.6$ & $17.4$ & $34.7$ & $13.0$  & $4.9$ & $50.3$ & $5.5$ & $5.8$ & $1.1$ \\
Maximum Softmax    & & & & $30.1$ & $33.2$ & $14.2$ & $\mathbf{62.2}$ & $10.3$  & $15.7$ & $16.6$ & $19.7$ & $15.9$ & $6.3$ \\
Entropy            & & & & $52.0$ & $30.0$ & $40.4$ & $53.8$ & $42.4$  & $\mathbf{20.6}$ & $\mathbf{16.3}$ & $\mathbf{21.4}$ & $\mathbf{19.5}$ & $\mathbf{10.4}$ \\ 
$\mathrm{PGN}_{\mathit{uni}}^{p=0.5}$ (ours) & & & & $\mathbf{69.3}$ & $\mathbf{9.8}$ & $\mathbf{50.0}$ & $44.8$ & $\mathbf{45.4}$  & $16.5$ & $19.7$ & $19.5$ & $14.8$ & $7.4$ \rule{0mm}{3.5mm}\\
\bottomrule
\end{tabular} }
\label{tab:ood_obstacle}
\end{table*}
\begin{table*}[t]
\caption{OoD segmentation benchmark results for the Fishyscapes LostAndFound and the RoadAnomaly21 dataset.}
\centering
\resizebox{0.98\linewidth}{!}{
\begin{tabular}{c ccc cc ccc cc ccc}
\toprule 
\multicolumn{1}{c}{} & \multirow{2}{*}{\rotatebox{90}{\footnotesize OoD data}} & \multirow{2}{*}{\rotatebox{90}{\footnotesize OoD arch}} & \multirow{2}{*}{\rotatebox{90}{\footnotesize adversarial}} & \multicolumn{5}{c}{Fishyscapes LostAndFound} & \multicolumn{5}{c}{RoadAnomaly21} \\[7pt]
\cmidrule(r){5-9} \cmidrule(r){9-14}
 \multicolumn{4}{c}{} & AuPRC $\uparrow$ & FPR$_{95}$ $\downarrow$ & $\overline{\text{sIoU}}$ $\uparrow$ & $\overline{\text{PPV}}$ $\uparrow$ & $\overline{F_1}$ $\uparrow$ & AuPRC $\uparrow$ & FPR$_{95}$ $\downarrow$ & $\overline{\text{sIoU}}$ $\uparrow$ & $\overline{\text{PPV}}$ $\uparrow$ & $\overline{F_1}$ $\uparrow$ \\[3pt]
\midrule 
Void Classifier    & $\checkmark$ & & & $11.7$ & $15.3$ & $9.2$ & $39.1$ & $14.9$   & $36.6$ & $63.5$ & $21.1$ & $22.1$ & $6.5$ \\
Maximized Entropy  & $\checkmark$ & ($\checkmark$) & & $44.3$ & $37.7$ & $21.1$ & $\mathbf{48.6}$ & $30.0$  & $\mathbf{85.5}$ & $\mathbf{15.0}$ & $\mathbf{49.2}$ & $\mathbf{39.5}$ & $\mathbf{28.7}$ \\
SynBoost           & $\checkmark$ & $\checkmark$ & & $\mathbf{64.9}$ & $30.9$ & $\mathbf{27.9}$ & $\mathbf{48.6}$ & $\mathbf{38.0}$  & $56.4$ & $61.9$ & $34.7$ & $17.8$ & $10.0$ \\
Image Resynthesis  & & $\checkmark$ & & $5.1$ & $29.8$ & $5.1$ & $12.6$ & $4.1$     & $52.3$ & $25.9$ & $39.7$ & $11.0$ & $12.5$ \\
Embedding Density  & & $\checkmark$ & & $8.9$ & $42.2$ & $5.9$ & $10.8$ & $4.9$     & $37.5$ & $70.8$ & $33.9$ & $20.5$ & $7.9$ \\
ODIN               & & & $\checkmark$ & $15.5$ & $38.4$ & $9.9$ & $21.9$ & $9.7$    & $33.1$ & $71.7$ & $19.5$ & $17.9$ & $5.2$ \\
Mahalanobis        & & & $\checkmark$ & $32.9$ & $\mathbf{8.7}$ & $19.6$ & $29.4$ & $19.2$   & $20.0$ & $87.0$ & $14.8$ & $10.2$ & $2.7$ \\
\midrule
Ensemble           & & ($\checkmark$) & & $0.3$ & $90.4$ & $3.1$ & $1.1$ & $0.4$      & $17.7$ & $91.1$ & $16.4$ & $20.8$ & $3.4$ \\
MC Dropout         & & ($\checkmark$) & & $14.4$ & $47.8$ & $4.8$ & $18.1$ & $4.3$    & $28.9$ & $69.5$ & $20.5$ & $17.3$ & $4.3$  \\ 
Maximum Softmax    & & & & $5.6$ & $40.5$ & $3.5$ & $9.5$ & $1.8$      & $28.0$ & $72.1$ & $15.5$ & $15.3$ & $5.4$ \\
Entropy            & & & & $14.0$ & $39.3$ & $8.0$ & $17.5$ & $7.7$    & $31.6$ & $71.9$ & $15.7$ & $18.4$ & $4.2$ \\
$\mathrm{PGN}_{\mathit{uni}}^{p=0.5}$ (ours)  & & & & $\mathbf{26.9}$ & $\mathbf{36.6}$ & $\mathbf{14.8}$ & $\mathbf{29.6}$ & $\mathbf{16.5}$  & $\mathbf{42.8}$ & $\mathbf{56.4}$ & $\mathbf{25.8}$ & $\mathbf{21.8}$ & $\mathbf{9.7}$ \rule{0mm}{3.5mm}\\
\bottomrule
\end{tabular} }
\label{tab:ood_anomaly}
\end{table*}
\begin{table}[t]
\caption{Runtime measurements in seconds per frame for each method; standard deviations taken over samples in the Cityscapes validation dataset.}
\centering
\resizebox{0.4\textwidth}{!}{
    \begin{tabular}{ c c c } \toprule
    sec.\ per frame $\downarrow$ & WideResNet & SEResNeXt \\\midrule
    Softmax     & \(3.02 \pm 0.18\) & \(1.08 \pm 0.04\) \\
    MC Dropout  & \(7.52 \pm 0.35\) & \(2.31 \pm 0.10\) \\
    \(\pgnoh + \pgnuni\) (ours) & \(3.05 \pm 0.16\) & \(1.09 \pm 0.05\) \\
     \bottomrule
    \end{tabular}
}
\label{tab:runtime}
\end{table}
To reduce the number of FP predictions and to estimate the prediction quality, we use meta classification and meta regression introduced by~\cite{Rottmann2018}. 
As input for the post-processing models, the authors use information from the network's softmax output which characterize uncertainty and geometry of a given segment like the segment size. 
The degree of randomness in semantic segmentation prediction is quantified by pixel-wise quantities, like entropy and probability margin. 
Segment-wise features are generated from these quantities via average pooling. 
These MetaSeg features used in~\cite{Rottmann2018} serve as a baseline in our tests.

Similarly, we construct segment-wise features from our pixel-wise gradient $p$-norms for $p \in \{ 0.1, 0.3, 0.5, 1, 2 \}$. 
We compute the mean and the variance of these pixel-wise values over a given segment. 
These features per segment are considered also for the inner and the boundary since the gradient scores may be higher on the boundary of a segment, see Figure~\ref{fig:seg_ood} (bottom). 
The inner of the segment consists of all pixels whose eight neighboring pixels are also elements of the same segment. 
Furthermore, we define some relative mean and variance features over the inner and boundary which characterizes the degree of fractality.
These hand-crafted quantities form a structured dataset where the columns correspond to features and the rows to predicted segments which serve as input to the post-processing models. 
Details on the construction of these features can be found in Appendix~\ref{sec:app_B}. 

We determine the prediction accuracy of semantic segmentation networks with respect to the ground truth via the segment-wise intersection over union ($\IoU$,~\cite{Jaccard1912}). 
On the one hand, we perform the direct prediction of the $\IoU$ (meta regression) which serves as prediction quality estimate. 
On the other hand, we discriminate between $\IoU = 0$ (FP) and $\IoU > 0$ (true positive) (meta classification). 
We use linear classification and regression models.

For the evaluation, we use $\auroc$ (area under the receiver operating characteristic) for meta classification and determination coefficient $R^2$ for meta regression. 
As baselines, we employ an ensemble, MC Dropout, maximum softmax probability, entropy and the MetaSeg framework.
A comparison of these methods and our approach for the Cityscapes dataset is given in Figure~\ref{fig:metaseg}. 
We outperform all baselines 
with the only exception of meta classification for the SEResNeXt backbone where MetaSeg achieves a marginal $0.41$ pp higher $\auroc$ value than ours. 
Such a post-processing model captures all the information which is contained in the network's output. 
Therefore, matching the MetaSeg performance by a gradient heatmap is a substantial achievement for a predictive uncertainty method.
Moreover, we enhance the MetaSeg performance for both networks and both tasks combining the MetaSeg features with $\text{PGN}$ by up to $1.02$ pp for meta classification and up to $3.98$ pp for meta regression.
This indicates that gradient features contain information which is partially orthogonal to the information contained in the softmax output.
Especially, the highest $\auroc$ value of $93.31\%$ achieved for the WideResNet backbone, shows the capability of our approach to estimate the prediction quality and filter out FP predictions on the segment-level for in-distribution data.
%
%
\paragraph{OoD Segmentation.} 
Figure~\ref{fig:heatmaps} shows segmentation predictions of the pre-trained DeepLabv3+ network with the WideResNet38 backbone together with its corresponding $\pgnuni^{p=0.5}$-heatmap.
The two panels on the left show an in-distribution prediction on Cityscapes where uncertainty is mainly concentrated around segmentation boundaries which are always subject to high prediction uncertainty.
Moreover, we see some false predictions in the far distance around the street crossing which can be found as a region of high gradient norm in the heatmap.
In the two panels to the right, we see an OoD prediction from the RoadAnomaly21 dataset of a sloth crossing the street which is classified as part vegetation, terrain and person.
The outline of the segmented sloth can be seen brightly in the gradient norm heatmap to the right indicating clear separation.

Our results in OoD segmentation are based on the evaluation protocol of the official SegmentMeIfYouCan benchmark~\cite{Chan2021_1}.
Evaluation on the pixel-level involves the threshold-independent area under precision-recall curve ($\text{AuPRC}$) and the false positive rate at the point of $0.95$ true positive rate ($\text{FPR}_{95}$).
The latter constitutes an interpretable choice of operating point for each method where a minimum true positive fraction is dictated.
On segment-level, an adjusted version of the mean intersection over union ($\text{sIoU}$) representing the accuracy of the segmentation obtained by thresholding at a particular point, positive predictive value ($\text{PPV}$ or precision) playing the role of binary instance-wise accuracy and the $F_1$-score.
The latter segment-wise metrics are averaged over thresholds between $0.25$ and $0.75$ with a step size of $0.05$ leading to the quantities $\overline{\text{sIoU}}$, $\overline{\text{PPV}}$ and $\overline{F_1}$.

Here, we compare the gradient scores of \(\pgnuni\) as an OoD score against all other methods from the benchmark which provide evaluation results on all given datasets. 
Note, we include a table of methods with incomplete results in Appendix~\ref{sec:app_C}. 
We restrict ourselves to $\pgnuni$ since we suspect it performs especially well in OoD segmentation as explained in Section~\ref{sec:method}.
The results are based on evaluation files submitted to the public benchmark and are, therefore, deterministic.
As baselines, we also include the same methods as for error detection before since these constitute a fitting comparison. 
Note, that the standard entropy baseline is not featured in the official leaderboard, so we report our own results obtained by the softmax entropy with the WideResNet backbone which performed better.

The results verified by the official benchmark are compiled in Table~\ref{tab:ood_obstacle} and Table~\ref{tab:ood_anomaly}.
We separate both tables into two halves. The bottom half contains methods with similar requirements as our method while the top half contains the stronger OoD segmentation methods which may have heavy additional requirements such as auxiliary data, architectural changes, retraining or the computation of adversarial examples.
Note, the best performance value of both halves of the table is marked.
In the lower part, a comparison with methods which have similarly low additional requirements is shown. 
We mark deep ensembles and MC dropout as requiring architectural changes since they technically require re-training in order to serve as numerical approximations of a Bayesian neural network. 
$\text{PGN}$ performs among the strongest methods, showing superior performance on LostAndFound test-NoKnown, Fishyscapes and the RoadAnomaly dataset.
While not clearly superior to the Entropy baseline on RoadObstacles, $\text{PGN}$ still yields stronger performance than the deep ensemble and MC dropout baselines.
Especially in Table~\ref{tab:ood_anomaly}, the methods in the lower part are significantly weaker w.r.t.\ most of the computed metrics.

The upper part of the table provides results for other OoD segmentation methods utilizing adversarial samples (ODIN \cite{Liang2018} and Mahalanobis \cite{Lee2018}), OoD data (Void Classifier \cite{Blum2019_1}, Maximized Entropy \cite{Chan2021} and SynBoost \cite{Biase2021}) or auxiliary models (SynBoost, Image Resynthesis \cite{Lis2019} and Embedding Density \cite{Blum2019_1}). 
Note, that the Maximized Entropy method does not modify the network architecture, but uses another loss function requiring retraining.
In several cases, we find that our method is competitive with some of the stronger methods. 
We outperform the two adversarial-based methods for the LostAndFound as well as the RoadAnomaly21 dataset. 
In detail, we obtain AuPRC values up to $22.8$ pp higher on segment-level and $\overline{F_1}$ values values up to $24.8$ pp higher on pixel-level. 
For the other two datasets we achieve similar results. 
Furthermore, we beat the Void Classifier method, that uses OoD data during training, in most cases. 
We improve the AuPRC metric by up to $64.5$ pp and the $\overline{\text{sIoU}}$ metric by up to $48.2$ pp, both for the LostAndFound dataset. 
In addition, our gradient norm outperforms in many cases the Image Resynthesis as well as the Embedding Density approach which are based on an external reconstruction model followed by a discrepancy network and normalizing flows, respectively. 
Summing up, we have shown superior OoD segmentation performance in comparison to the other uncertainty based methods and outperform some of the more complex approaches (using OoD data, adversarial samples or generative models). 
%
%
\paragraph{Computational Runtime.} 
Lastly, we demonstrate the computational efficiency of our method and show runtime measurements for the network forward pass (``Softmax''), MC dropout (\(25\) samples) and computing both, \(\pgnoh\) and \(\pgnuni\), in Table~\ref{tab:runtime}.
While sampling MC dropout requires over twice the time per frame for both backbone networks, the computation of \(\pgnoh + \pgnuni\) only leads to a marginal computational overhead of around \(1 \%\) due to consisting only of tensor multiplications.
Note, that the entropy baseline requires roughly the same compute as Softmax and deep ensembles tend to be slower than MC dropout due to partial parallel computation.
The measurements were each made on a single Nvidia Quadro P6000 GPU.
%
%
\paragraph{Limitations.} 
We declare that our method is merely on-par with other uncertainty quantification methods for pixel-level considerations but outperforming well-known methods such as MC Dropout. 
For segment-wise error detection, our approach showed improved results. 
Moreover, there have been some submissions to the SegmentMeIfYouCan benchmark (requiring OoD data, re-training or complex auxiliary models) which outperform our method. 
However, a direct application of the methods is barely possible if suitable OoD data have to pass into the training or auxiliary models are used in combination with the segmentation network, which also increases the runtime during application, in comparison to our simple and flexible method. 
Note, our approach has one light architectural restrictions, i.e., the final layer has to be a convolution, but this is in general common for segmentation models.
%
%
%
\section{\uppercase{Conclusion}}\label{sec:conc}
In this work, we presented an efficient method of computing gradient uncertainty scores for a wide class of deep semantic segmentation models.
Moreover, we appreciate a low computational cost associated with them.
Our experiments show that large gradient norms obtained by our method statistically correspond to erroneous predictions already on the pixel-level and can be normalized such that they yield similarly calibrated confidence measures as the maximum softmax score of the model.
On a segment-level our method shows considerable improvement in terms of error detection.
Gradient scores can be utilized to segment out-of-distribution objects significantly better than sampling- or any other output-based method on the SegmentMeIfYouCan benchmark and has competitive results with a variety of methods, in several cases clearly outperforming all of them while coming at negligible computational overhead.
We hope that our contributions in this work and the insights into the computation of pixel-wise gradient feedback for segmentation models will inspire future work in uncertainty quantification and pixel-wise loss analysis.

\section*{ACKNOWLEDGEMENTS} 
We thank Hanno Gottschalk and Matthias Rottmann for discussion and useful advice.
This work is supported by the Ministry of Culture and Science of the German state of North Rhine-Westphalia as part of the KI-Starter research funding program.
The research leading to these results is funded by the German Federal Ministry for Economic Affairs and Climate Action within the project “KI Delta Learning“ (grant no.\ 19A19013Q). The authors would like to thank the consortium for the successful cooperation.
The authors gratefully acknowledge the Gauss Centre for Supercomputing e.V. \url{www.gauss-centre.eu} for funding this project by providing computing time through the John von Neumann Institute for Computing (NIC) on the GCS Supercomputer JUWELS at Jülich Supercomputing Centre (JSC).

\bibliographystyle{apalike}
{\small
\bibliography{example}}

\newpage
\appendix
\section*{APPENDIX}
%
%
\section{Computational Derivations and Implementation Details}\label{sec:app_A}

\subsection{Computations}
In this work we assume that the final activation in the segmentation network is given by a softmax activation of categorical features $\phi$ in the last layer
\begin{equation}
    \varSigma^j (\phi) = \frac{\e{\phi^j}}{\sum_{i = 1}^C \e{\phi^i}}
\end{equation}
where $C$ is the number of classes and $\phi = \phi(\theta)$ is now dependent on a set of parameters $\theta$.
In order to compute the gradients of the categorical cross entropy at pixel $(a,b)$ given any auxiliary label $y$
\begin{equation}
    \mathcal{L}_{ab}(\phi(\theta) \| y) = - \sum_{j = 1}^C y_{ab, j} \log(\varSigma^j(\phi_{ab}(\theta))),
\end{equation}
we require the derivative of the softmax function
\begin{align}
    \begin{split}
        \nabla_{\! \theta} \varSigma^j(\phi(\theta)) 
        =&
        \frac{\e{\phi^j} \nabla_{\! \theta} \phi^j}{\sum_{i = 1}^C \e{\phi^i}} - \frac{\e{\phi^j} \sum_{k = 1}^C \e{\phi^k} \nabla_{\! \theta} \phi^k}{\left( \sum_{i = 1}^C \e{\phi^i} \right)^2} \\
        =& 
        \varSigma^j(\phi) \cdot \nabla_{\! \theta} \phi^j \sum_{k = 1}^C \varSigma^k(\phi) \\
        &- \varSigma^j(\phi) \sum_{k = 1}^C \varSigma^k(\phi) \cdot \nabla_{\! \theta} \phi^k \\
        =& 
        \varSigma^j(\phi) \left( \sum_{k = 1}^C \varSigma^k(\phi) \cdot \nabla_{\! \theta} \phi^j - \varSigma^k \nabla_{\! \theta} \phi^k\right) \\
        =&
        \varSigma^j(\phi) \sum_{k = 1}^C \varSigma^k(\phi) (\delta_{kj} - 1) \nabla_{\! \theta} \phi^k.
    \end{split}
\end{align}
Note, that the auxiliary label $y \in [0, 1]^C$ may be anything, e.g., actual ground truth information about pixel $(a,b)$, the predicted probability distribution at that location, the one-hot encoded prediction or a uniform class distribution.
In the following, we regard $y$ to be independent of $\phi$.
The gradient of the cross entropy loss is
\begin{align}
    \begin{split}
        \nabla_{\! \theta} \mathcal{L}_{ab}(\phi(\theta) \| y)
        =& 
        - \sum_{j = 1}^C y_{ab,j} \frac{1}{\varSigma^j(\phi_{ab})} \nabla_{\! \theta} \varSigma^j(\phi_{ab}) \\
        =&
        \sum_{j = 1}^C \sum_{k = 1}^C (y_{ab,j} \varSigma^k(\phi_{ab}) \\
        & - y_j \varSigma^k(\phi_{ab}) \delta_{kj}) \cdot \nabla_{\! \theta} \phi^k_{ab}(\theta) 
        \\
        =&
        \sum_{k = 1}^C \varSigma^k(\phi_{ab}) (1 - y_k) \cdot \nabla_{\! \theta} \phi^k_{ab}(\theta) .
    \end{split}
\end{align}
The feature maps are the result of convolution $\mathcal{C}_{K, \beta}$ against a filter bank $K \in \R^{\kappa_{\mathrm{in}} \times \kappa_{\mathrm{out}} \times (2 s + 1) \times (2 s + 1)}$ and addition of a bias vector $\beta \in \R^{\kappa_\mathrm{out}}$.
Here, $\kappa_\mathrm{in}$ and $\kappa_\mathrm{out}$ denote the number of in- and out-going channels, respectively.
$K$ has parameters $(K_e^h)_{fg}$ where $e$ and $h$ indicate in- and out-going channels respectively and $f$ and $g$ index the spatial filter position.
In particular, we obtain the value $\phi_{ab}^d$ at pixel location $(a, b)$ in channel $d$ by
\begin{align}
    \begin{split}
    \phi_{ab}^d(K, \beta, \psi) 
    =& 
    [\mathcal{C}_{K, \beta}(\psi)]_{ab}^d
    = 
    (K \ast \psi)_{ab}^d + \beta^d \\
    =&
    \sum_{c = 1}^{\kappa_\mathrm{in}} \sum_{q=-s}^s \sum_{r = -s}^s (K_c^d)_{qr} \psi^c_{a + q, b + r} + \beta^d.
    \end{split}
\end{align}
Here, we assume the filters of size $(2 s + 1) \times (2 s + 1)$ indexed symmetrically around their center and $\psi$ is the feature map pre-convolution.
Note, also that the bias is constant across the indices $a$ and $b$.
Taking explicit derivatives of this expression with respect to one of the parameters in $K$ yields
\begin{align}
    \begin{split}
    \frac{\partial (K\ast \phi)_{ab}^d}{\partial (K_e^h)_{fg}} 
    =&
    \sum_{c = 1}^{\kappa_\mathrm{in}} \sum_{q,r = -s}^s \frac{\partial (K_c^d)_{qr}}{\partial (K_e^h)_{fg}} \, \psi_{a+q, b+r}^c \\
    =&
    \sum_{c = 1}^{\kappa_\mathrm{in}} \sum_{q,r = -s}^s \delta^{dh} \delta_{ce} \delta_{pf} \delta_{qg} \psi_{a+q, b+r}^c \\
    =& \delta^{dh}\psi_{a+f, b+g}^e.
    \end{split}
\end{align}
When utilizing the predicted one-hot vector (self-learning gradient), i.e., $y^\mathit{oh}_k = \delta_{k \hat{c}}$ and using the fact that the last-layer convolution is $(1 \times 1)$, we obtain
\begin{align}
    \begin{split}
    \frac{\partial}{\partial K_e^f} \mathcal{L}_{ab}
    =&
    \sum_{k = 1}^C \varSigma^k(\phi_{ab}) (1 - \delta_{k \hat{c}}) \frac{\partial \phi_{ab}^k}{\partial K_e^f} \\
    =&
    \varSigma^f(\phi_{ab}) (1 - \delta_{f \hat{c}}) \psi_{ab}^e
    \end{split}
\end{align}
while the uniform categorical label $y_k = \tfrac{1}{C}$ yields
\begin{align}
    \begin{split}
    \frac{\partial}{\partial K_e^f} \mathcal{L}_{ab}
    =&
    \sum_{k = 1}^C \varSigma^k(\phi_{ab}) \left(1 - \frac{1}{C}\right) \cdot \nabla_{\! \theta} \phi^k_{ab}(\theta) \\
    =&
    \frac{C - 1}{C} \varSigma^f(\phi_{ab}) \psi_{ab}^e.
    \end{split}
\end{align}

\subsection{Computing Deeper Gradients via Explicit Backpropagation}\label{sec:deep_grad}
While for the last-layer gradients the computation from above simplifies significantly due to the fact that pixel-wise gradients only depend on the feature map values at the same pixel-location.
Deeper layers are usually not $(1 \times 1)$, so the forward pass couples feature map values over a larger receptive field, e.g., $(3 \times 3)$.
However, gradients for the second-to-last layer can still be computed with comparably small effort by extracting feature map patches via the unfold function.
Here, we consider a DeepLabv3+ implementation$^1$\let\thefootnote\relax\footnotetext{$^1$\url{https://github.com/NVIDIA/semantic-segmentation/tree/sdcnet}} \cite{Zhu2019} for which the feature maps depend in the following way on the weights of the second-to-last layer $T - 1$ (where $T$ indicates the last layer):
\begin{equation}
    \phi(K_{T - 1}) = \mathcal{C}_{K_T, \beta_T} \circ \mathsf{ReLU} \circ \mathsf{BN}_T \circ \mathcal{C}_{K_{T - 1}, \beta_{T - 1}}(\psi_{T - 1}),
\end{equation}
where $\mathsf{BN}_T$ is a batch normalization layer and $\psi_{T - 1}$ are the features prior to the convolution $\mathcal{C}_{K_{T - 1}, \beta_{T - 1}}$ in question.
With fully expanded indices, this amounts to
\begin{align}
    \begin{split}
    \phi_{ab}^d =& \sum_{k_T} K_{k_T}^d \mathsf{ReLU} \Biggl( \mathsf{BN}_T \Biggl[ \sum_{k_{T-1}} \sum_{q,r} \left((K_{T-1})_{k_{T-1}}^{k_T}\right)_{qr} \\
    & (\psi_{T-1})_{a+q, b+r}^{k_{T-1}} + \beta_{T-1}^{k_T}\biggr]\Biggr) + \beta_{T}^d
    \end{split}
\end{align}
with $k_T$, $k_{T-1}$ indexing the respective amount of channels and $q, r$ indexing the filter coordinates of $K_{T-1}$.
Note, that we still assume here that the last convolutional layer has spatial extent 0 into both directions, so $K_T \in \R^{\kappa_\mathrm{in} \times C \times 1 \times 1}$.
The chain rule for this forward pass is then
\begin{align}\label{eq: derivation of deep gradient formula}
    \begin{split}
        & \frac{\partial \phi_{ab}^d}{\partial ((K_{T-1})_e^f)_{gh}}
        =
        \sum_{k_T} K_{k_T}^d \mathsf{ReLU}'(\cdot) \cdot \mathsf{BN}_T' \\
        & \ \ \ \ \ \ \cdot \left(\sum_{k_{T-1}} \sum_{q,r} \frac{\partial ((K_{T-1})_{k_{T-1}}^{k_T})_{qr}}{\partial ((K_{T-1})_e^f)_{gh}} (\psi_{T-1})_{a+q, b+r}^{k_{T-1}}\right) \\
        & \ \ \ \ =
        \sum_{k_T} K_{k_T}^d \mathsf{ReLU}'(\cdot) \cdot \mathsf{BN}'_T \\
        & \ \ \ \ \ \ \cdot \sum_{k_{T-1}} \sum_{q,r} \delta^{k_T f} \delta_{k_{T-1}e} \delta_{gp} \delta_{hq} (\psi_{T-1})_{a+q, b+r}^{k_{T-1}} \\
        & \ \ \ \ =
        K_f^d (\mathsf{ReLU}')_{ab}^f \cdot (\mathsf{BN}'_T)_{ab}^f \cdot (\psi_{T-1})_{a+g, b+h}^e .
    \end{split}
\end{align}
The term with the derivative of the $\mathsf{ReLU}$ activation is simply the Heaviside function evaluated at the features pre-activation which have been computed in the forward pass anyway (the discontinuity at zero has vanishing probability of an evaluation).
The derivative of the batch normalization layer is multiplication by the linear batch norm parameter which is applied channel-wise.
The running indices $g$ and $h$ only apply to the last factor which can be computationally treated by extracting $(3 \times 3)$-patches from $\psi_{T-1}$ using the unfold operation.
In computing the norm of $\nabla_{K_{T-1}} \phi_{ab}^d$ with respect to the indices $e$, $f$, $g$ and $h$, we note that the expression in eq.~\eqref{eq: derivation of deep gradient formula} is a product of two tensors $S_f \cdot \Psi_{gh}^e$ for each pixel $(a,b)$.
$L^p$-norms of such tensors $T_{ij} = S_i \Psi_j$ factorize which makes their computation feasible:
\begin{align}
    \begin{split}
    \|T\|_p =& \left( \sum_{i, j} |T_{ij}|^p \right)^{\frac{1}{p}}
    = \left(\sum_{i, j} |S_i|^p |\Psi_j|^p\right)^{\frac{1}{p}} \\
    =& \left(\sum_{i}\left[|S_i|^p \sum_j |\Psi_j|^p\right]\right)^{\frac{1}{p}}
    = \|S\|_p \|\Psi\|_p.
    \end{split}
\end{align}
%
%
\section{Details on the Feature Construction for Segment-wise Prediction Quality Estimation}\label{sec:app_B}
The tasks of meta classification (false positive detection) and meta regression (prediction quality estimation) based on uncertainty measures extracted from the network's softmax output were introduced in \cite{Rottmann2018}. The neural network provides for each pixel $z$ given an input image $x$ a probability distribution $f_{z}(y|x)$ over a label space $C = \{y_{1}, \ldots, y_{c} \}$, with $y \in C$. The degree of randomness in the semantic segmentation prediction $f_{z}(y|x)$ is quantified by pixel-wise dispersion measures, like entropy 
\begin{equation}
    E_z(x) =-\frac{1}{\log(c)}\sum_{y\in \mathcal{C}}f_z(y|x)\log f_z(y|x) 
\end{equation} 
and probability margin
\begin{equation}
    M_z(x) = 1 - f_z(\hat y_z(x)|x)  + \max_{y\in\mathcal{C}\setminus\{\hat y_z(x)\}} f_z(y|x) 
\end{equation}
where 
\begin{equation}
    \hat y_z(x)=\argmax_{y\in\mathcal{C}}f_z(y|x) 
\end{equation}
is the predicted class for each pixel $z$.
To obtain segment-wise features from these dispersion measures which characterize uncertainty, they are aggregated over segments via average pooling obtaining mean dispersions $\mu E$ and $\mu M$. As erroneous or poor predictions oftentimes have fractal segment shapes, it is distinguished between the inner of the segment (consisting of all pixels whose eight neighboring pixels are also elements of this segment) and the boundary. This results in segment size $S$ and mean dispersion features per segment also for the inner (\emph{in}) and the boundary (\emph{bd}). Furthermore, relative segment sizes $\tilde S = S/S_{bd}$ and $\tilde S_{in} = S_{in}/S_{bd}$ as well as relative mean dispersions $\mu 
 \tilde D = \mu D \tilde S$ and $\mu \tilde D_{in} = \mu D_{in} \tilde S_{in}$ where $D \in \{E,M\}$ are defined. Moreover, the mean class probabilities $P(y)$ for each class $y \in C$ are added resulting in the set of hand-crafted features 
\begin{align}  
    \begin{split}
    & \{ S, S_{in}, S_{bd}, \tilde S, \tilde S_{in} \} \cup \{ P(y) \, : \, y=1,\ldots,c \} \\
    & \cup \{ \mu D, \mu D_{in}, \mu D_{bd}, \mu \tilde D, \mu \tilde D_{in} \, : \, D \in \{E,M\} \} \, . 
    \end{split}
\end{align}
These features serve as a baseline in our tests.

The computed gradients in Section~\ref{sec:method} 
with applied $p$-norm, $p \in \{ 0.1, 0.3, 0.5, 1, 2 \}$, are denoted by $G_{z}^{p=\#}(x)$, $\# \in \{ 0.1, 0.3, 0.5, 1, 2 \}$,  for image $x$. Similar to the dispersion measures, we compute the mean and additionally the variance of these pixel-wise values of a given segment to obtain $\mu G^{p=\#}$ and $vG^{p=\#}$, respectively. Since the gradient uncertainties may be higher on the boundary of a segment, the mean and variance features per segment are considered also for the inner and the boundary. Furthermore, we define relative mean and variance features to quantify the degree of fractality. This results in the following gradient features
\begin{align}\label{eq:meta_grads}
    \begin{split}
     & \{ \mu G^{p=\#}, \mu G_{in}^{p=\#}, \mu G_{bd}^{p=\#}, \mu \tilde G^{p=\#}, \mu \tilde G_{in}^{p=\#}, \\
     & vG^{p=\#}, vG_{in}^{p=\#}, vG_{bd}^{p=\#}, v \tilde G^{p=\#}, v \tilde G_{in}^{p=\#} : \\
     & \# \in \{ 0.1, 0.3, 0.5, 1, 2 \} \} \, . 
     \end{split}
\end{align}
Note, these features can be computed for the gradient scores obtained from the predictive one-hot ($\pgnoh$) and the uniform label ($\pgnuni$) as well as for gradients of deeper layers (see Section~\ref{sec:deep_grad}).
%
%
\section{Extended OoD Segmentation Results}\label{sec:app_C}
In Section~\ref{sec:exp_results}, 
we have compared our approach with different OoD segmentation methods that provide a complete evaluation on the four different benchmark datasets.
In Table~\ref{tab:app_ood_obstacle_extend} and Table~\ref{tab:app_ood_anomaly_extend}, we provide the comparison of our method with the other approaches part of the benchmark. The incomplete results render the comparison less fair. 
\begin{table*}[t]
\centering
\resizebox{\linewidth}{!}{
\begin{tabular}{l cc ccc cc ccc}
\toprule
\multicolumn{1}{c}{} & \multicolumn{5}{c}{LostAndFound test-NoKnown} & \multicolumn{5}{c}{RoadObstacle21} \\
\cmidrule(r){2-6} \cmidrule(r){7-11}
 & AuPRC $\uparrow$ & FPR$_{95}$ $\downarrow$ & $\overline{\text{sIoU}}$ $\uparrow$ & $\overline{\text{PPV}}$ $\uparrow$ & $\overline{F_1}$ $\uparrow$ & AuPRC $\uparrow$ & FPR$_{95}$ $\downarrow$ & $\overline{\text{sIoU}}$ $\uparrow$ & $\overline{\text{PPV}}$ $\uparrow$ & $\overline{F_1}$ $\uparrow$ \\
\midrule
PEBAL              & $-$ & $-$ & $-$ & $-$ & $-$                  & $5.0$ & $12.7$ & $29.9$ & $7.6$ & $5.5$ \\
DenseHybrid        & $78.7$ & $2.1$ & $46.9$ & $52.1$ & $52.3$    & $87.1$ & $0.2$ & $45.7$ & $50.1$ & $50.7$ \\
RPL+CoroCL    & $-$ & $-$ & $-$ & $-$ & $-$                 & $85.9$ & $0.6$ & $52.6$ & $56.7$ & $56.7$ \\
EAM    & $-$ & $-$ & $-$ & $-$ & $-$                 & $92.9$ & $0.5$ & $65.9$ & $76.5$ & $75.6$ \\
RbA    & $-$ & $-$ & $-$ & $-$ & $-$                 & $95.1$ & $0.1$ & $54.3$ & $59.1$ & $57.4$ \\
Mask2Anomaly    & $-$ & $-$ & $-$ & $-$ & $-$                 & $93.2$ & $0.2$ & $55.7$ & $75.4$ & $68.2$ \\
\arrayrulecolor{lightgray}
\midrule
Road Inpainting    & $82.9$ & $35.8$ & $49.2$ & $60.7$ & $52.3$   & $54.1$ & $47.1$ & $57.6$ & $39.5$ & $36.0$ \\
NFlowJS            & $89.3$ & $0.7$ & $54.6$ & $59.7$ & $61.8$    & $85.6$ & $0.4$ & $45.5$ & $49.5$ & $50.4$ \\
JSRNet             & $74.2$ & $6.6$ & $34.3$ & $45.9$ & $36.0$    & $28.1$ & $28.9$ & $18.6$ & $24.5$ & $11.0$ \\
FlowEneDet    & $79.8$ & $2.9$ & $43.8$ & $52.8$ & $48.1$                 & $73.7$ & $1.0$ & $42.6$ & $42.3$ & $40.0$ \\
DaCUP    & $81.4$ & $7.4$ & $38.3$ & $67.3$ & $51.1$                 & $81.5$ & $1.1$ & $37.7$ & $60.1$ & $46.0$ \\   
\midrule
$\mathrm{PGN}_{\mathit{oh}}^{p=0.5}$  & $64.9$ & $18.4$ & $48.3$ & $50.0$ & $46.9$  & $18.8$ & $14.8$ & $22.1$ & $16.5$ & $9.2$ \rule{0mm}{3.5mm}\\
$\mathrm{PGN}_{\mathit{uni}}^{p=0.5}$  & $69.3$ & $9.8$ & $50.0$ & $44.8$ & $45.4$  & $16.5$ & $19.7$ & $19.5$ & $14.8$ & $7.4$ \rule{0mm}{3.5mm}\\
\arrayrulecolor{black}
\bottomrule
\end{tabular} }
\caption{OoD segmentation results for the LostAndFound and the RoadObstacle21 dataset.}
\label{tab:app_ood_obstacle_extend}
\end{table*}
\begin{table*}[t]
\centering
\resizebox{\linewidth}{!}{
\begin{tabular}{l cc ccc cc ccc}
\toprule
\multicolumn{1}{c}{} & \multicolumn{5}{c}{Fishyscapes LostAndFound} & \multicolumn{5}{c}{RoadAnomaly21} \\
\cmidrule(r){2-6} \cmidrule(r){7-11}
 & AuPRC $\uparrow$ & FPR$_{95}$ $\downarrow$ & $\overline{\text{sIoU}}$ $\uparrow$ & $\overline{\text{PPV}}$ $\uparrow$ & $\overline{F_1}$ $\uparrow$ & AuPRC $\uparrow$ & FPR$_{95}$ $\downarrow$ & $\overline{\text{sIoU}}$ $\uparrow$ & $\overline{\text{PPV}}$ $\uparrow$ & $\overline{F_1}$ $\uparrow$ \\
\midrule
PEBAL              & $-$ & $-$ & $-$ & $-$ & $-$                 & $49.1$ & $40.8$  & $38.9$ & $27.2$ & $14.5$ \\
DenseHybrid        & $-$ & $-$ & $-$ & $-$ & $-$                 & $78.0$ & $9.8$  & $54.2$ & $24.1$ & $31.1$ \\
RPL+CoroCL    & $-$ & $-$ & $-$ & $-$ & $-$                 & $83.5$ & $11.7$ & $49.8$ & $30.0$ & $30.2$ \\
EAM    & $81.5$ & $-$ & $-$ & $-$ & $-$                 & $93.8$ & $4.1$ & $67.1$ & $53.8$ & $60.9$ \\
RbA    & $-$ & $-$ & $-$ & $-$ & $-$                 & $94.5$ & $4.6$ & $64.9$ & $47.5$ & $51.9$ \\
Mask2Anomaly    & $46.0$ & $4.4$ & $-$ & $-$ & $-$                 & $88.7$ & $14.6$ & $55.3$ & $51.7$ & $47.2$ \\
\arrayrulecolor{lightgray}
\midrule
NFlowJS            & $-$ & $-$ & $-$ & $-$ & $-$                 & $56.9$ & $34.7$ & $36.9$ & $18.0$ & $14.9$ \\
JSRNet             & $-$ & $-$ & $-$ & $-$ & $-$                 & $33.6$ & $43.9$ & $20.2$ & $29.3$ & $13.7$ \\
ObsNet             & $-$ & $-$ & $-$ & $-$ & $-$                 & $75.4$ & $26.7$ & $44.2$ & $52.6$ & $45.1$ \\
FlowEneDet    & $-$ & $-$ & $-$ & $-$ & $-$                 & $36.7$ & $77.8$ & $15.5$ & $16.8$ & $3.4$ \\
\arrayrulecolor{black}\midrule 
$\mathrm{PGN}_{\mathit{oh}}^{p=0.5}$  & $22.8$ & $35.5$ & $12.1$ & $27.3$ & $14.1$  & $39.3$ & $66.5$ & $23.1$ & $21.5$ & $7.8$ \rule{0mm}{3.5mm}\\
$\mathrm{PGN}_{\mathit{uni}}^{p=0.5}$  & $26.9$ & $36.6$ & $14.8$ & $29.6$ & $16.5$  & $42.8$ & $56.4$ & $25.8$ & $21.8$ & $9.7$ \rule{0mm}{3.5mm}\\
\arrayrulecolor{black}
\bottomrule
\end{tabular} }
\caption{OoD segmentation results for the Fishyscapes LostAndFound and the RoadAnomaly21 dataset.}
\label{tab:app_ood_anomaly_extend}
\end{table*}
In detail, the first block consists of approaches using OoD data, i.e., PEBAL \cite{Tian2022}, DenseHybrid \cite{Grcic2022}, RPL+CoroCL \cite{Liu2023}, EAM \cite{Grcic2023}, RbA \cite{Nayal2023} and Mask2Anomaly \cite{Rai2023}. 
The latter three methods are based on the strong vision transformer models resulting in comparatively high OoD segmentation performance.
The methods of the second block use complex auxiliary/generative models, namely Road Inpainting \cite{Lis2020}, NFlowJS \cite{Grcic2021}, JSRNet \cite{Vojir2021}, ObsNet \cite{Besnier2021}, FlowEneDet \cite{Gudovskiy2023} and DaCUP \cite{Vojir2023}. 
Our method shows performance values in the mid-range among the different methods for some metrics. However, these approaches require OoD data for retraining or are based on auxiliary or generative models, while our PGN module can be easily integrated into various semantic segmentation models during inference with minimal computation overhead.

Gradient scores perform perhaps the least convincingly on the RoadObstacle21 benchmark. 
We find a slight trend of $\pgnoh$ performing better in the Obstacle track which can be seen to be closer to in-distribution data in semantic segmentation for autonomous driving.
This connection would be consistent with our finding in actual in-distribution errors, while $\pgnuni$ performs better on the Anomaly track which is more clearly out-of-distribution for street scene recognition.
%
%
\section{Ablation: Different $p$-norms}\label{sec:app_D}
\begin{figure*}[t]
    \center
    \includegraphics[width=0.99\textwidth]{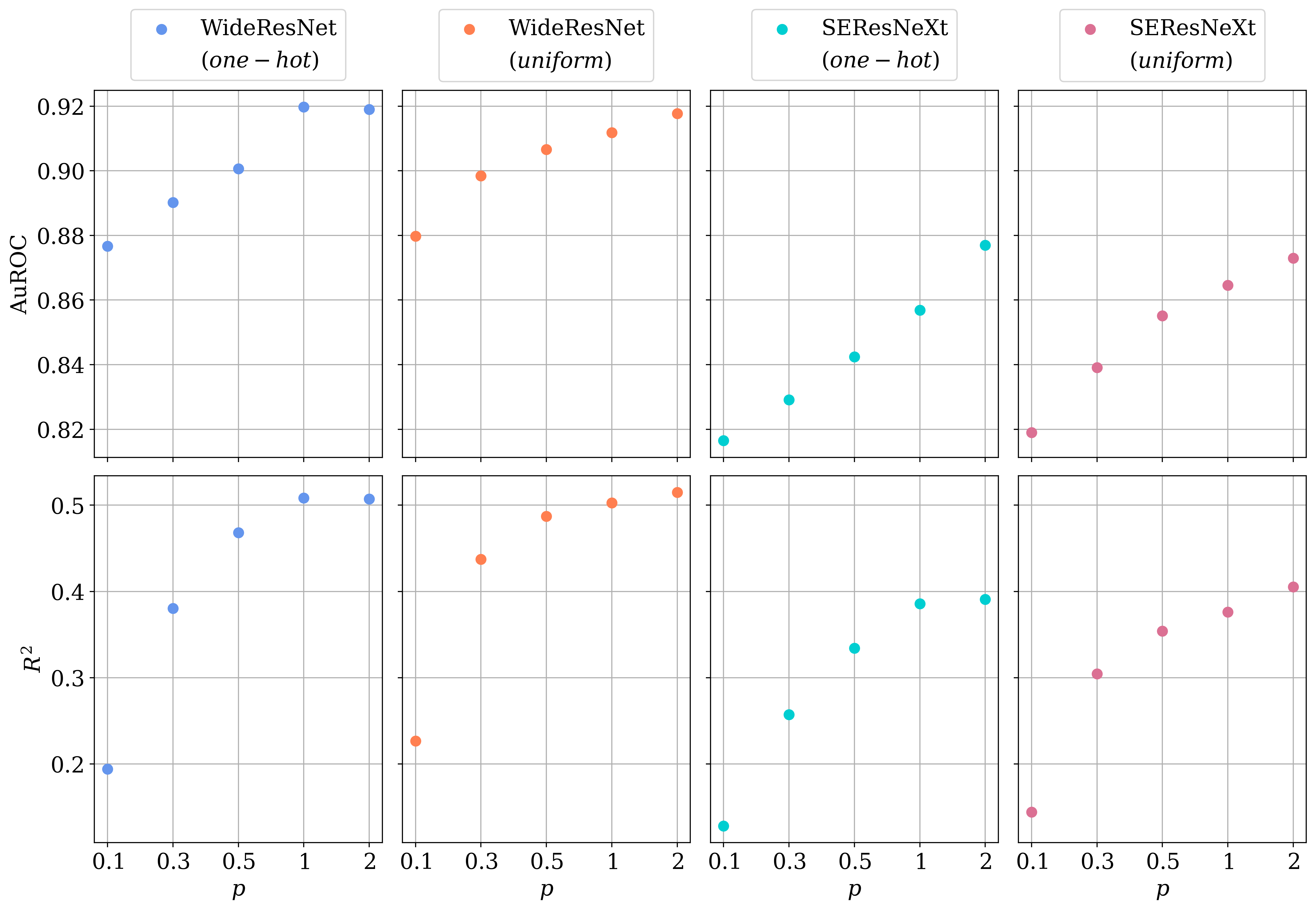}
    \caption{$\auroc$ and $R^2$ values for both backbone architectures applied to the Cityscapes dataset and the different $p$-norms.}
    \label{fig:app_metaseg}
\end{figure*}

\begin{figure*}
    \centering
    \resizebox{1.01\linewidth}{!}{
    \input{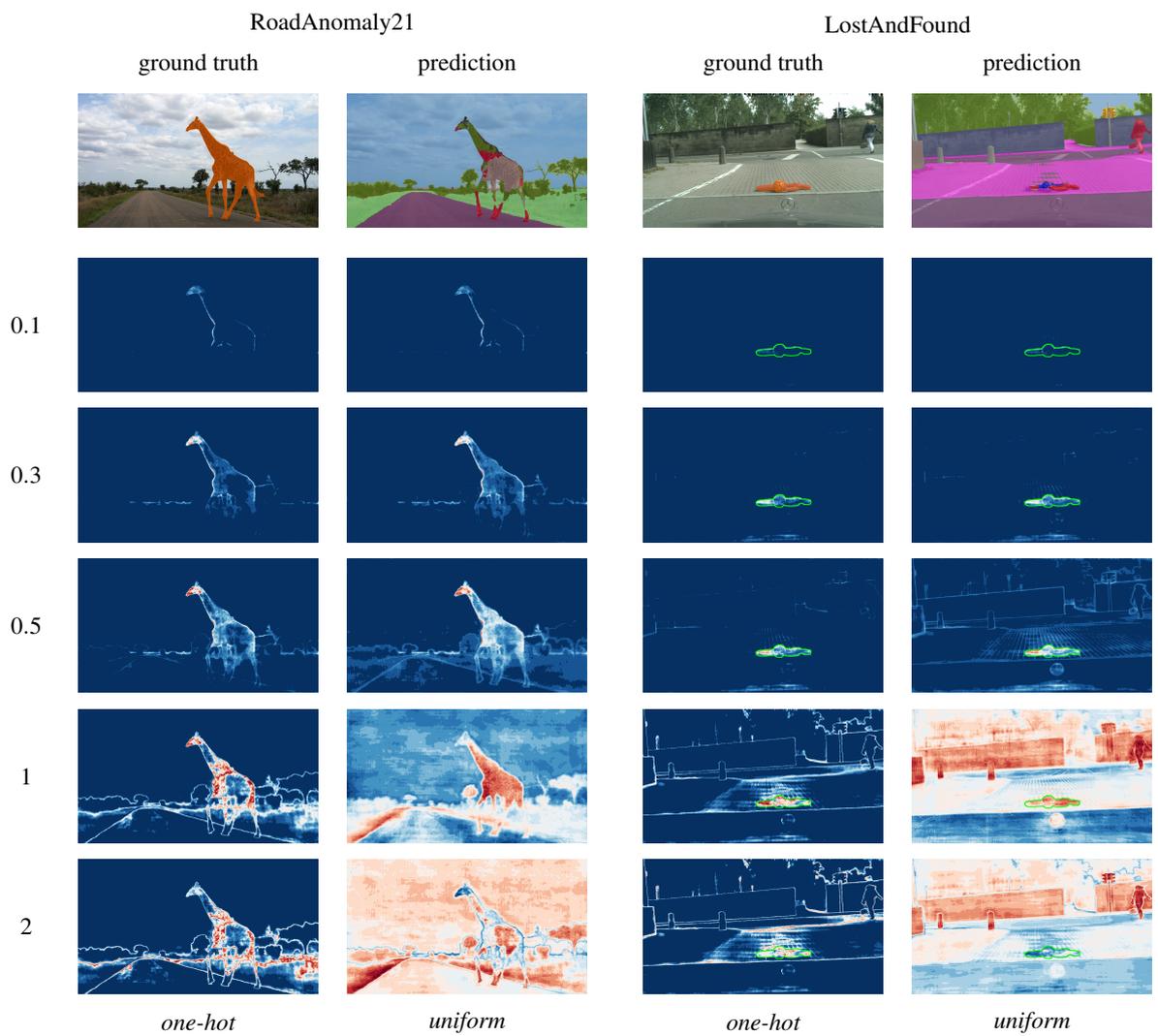} }
    \caption{Ground truth (labeled OoD object), semantic segmentation prediction and PGN heatmaps for different $p$-norms as well as for the predictive one-hot and the uniform label, respectively.}
    \label{fig:app_heatmaps}
\end{figure*}

In this section, we present an ablation study for different $p$-norms, $p \in \{ 0.1, 0.3, 0.5, 1, 2 \}$, that are applied to the calculated gradients of the last convolutional layer.
%
\paragraph{Pixel-wise Uncertainty Evaluation.}
The area under the sparsification error curve ($\ause$) is considered to access the correlation of uncertainty and prediction errors on pixel level and the expected calibration error ($\ece$) to assess the statistical reliability of the uncertainty measure. In Table~\ref{tab:app_eval_pix} the results for the different $p$-norms in terms of these two metrics are given.
\begin{table}[t]
\centering
\scalebox{0.8}{
\begin{tabular}{l c cc cc}
\toprule
\multicolumn{2}{c}{} & \multicolumn{2}{c}{last layer} & \multicolumn{2}{c}{second-to-last layer} \\
\cmidrule(r){3-4} \cmidrule(r){5-6}
 & $p$ & $\ece$ $\downarrow$ & $\ause$ $\downarrow$ & $\ece$ $\downarrow$ & $\ause$ $\downarrow$ \\
\midrule
 & $0.1$             & $0.0187$ & $0.0500$  & $0.0186$ & $\underline{0.0235}$ \\
Wide- & $0.3$        & $0.0183$ & $0.0712$  & $0.0125$ & $0.0286$ \\
ResNet & $0.5$       & $0.0163$ & $0.0508$  & $0.0059$ & $0.0280$ \\
(\emph{one-hot}) & $1$ & $\underline{0.0025}$ & $0.0307$  & $\underline{0.0027}$ & $0.0271$ \\
 & $2$                 & $\mathbf{0.0019}$ & $\underline{0.0268}$  & $\mathbf{0.0021}$ & $0.0265$ \\
\midrule
 & $0.1$               & $0.0186$ & $0.0784$  & $0.0096$ & $0.3347$ \\
Wide- & $0.3$          & $0.0163$ & $0.3426$  & $0.1846$ & $0.6746$ \\
ResNet & $0.5$         & $0.0241$ & $0.5857$  & $0.3385$ & $0.7520$ \\
(\emph{uniform}) & $1$ & $0.3762$ & $0.7424$  & $0.4345$ & $0.8028$ \\
 & $2$                 & $0.3868$ & $0.8104$  & $0.3989$ & $0.8253$ \\
\midrule
 & $0.1$               & $0.0347$ & $\mathbf{0.0201}$  & $0.0344$ & $\mathbf{0.0060}$ \\
SERes- & $0.3$         & $0.0336$ & $0.0386$  & $0.0290$ & $0.0353$ \\
NeXt & $0.5$           & $0.0305$ & $0.0399$  & $0.0214$ & $0.0379$ \\
(\emph{one-hot}) & $1$ & $0.0078$ & $0.0383$  & $0.0079$ & $0.0377$ \\
 & $2$                 & $0.0039$ & $0.0365$  & $0.0068$ & $0.0365$ \\
\midrule
 & $0.1$               & $0.0346$ & $0.0427$  & $0.0295$ & $0.1823$ \\
SERes- & $0.3$         & $0.0313$ & $0.2484$  & $0.1916$ & $0.4878$ \\
NeXt & $0.5$           & $0.0076$ & $0.5617$  & $0.3000$ & $0.6198$ \\
(\emph{uniform}) & $1$ & $0.3744$ & $0.7694$  & $0.4075$ & $0.7413$ \\
 & $2$                 & $0.4030$ & $0.8187$  & $0.4003$ & $0.8116$ \\
\bottomrule
\end{tabular} }
\caption{Pixel-wise uncertainty evaluation results for both backbone architectures and the Cityscapes dataset as well as for different $p$-norms and layers in terms of $\ece$ and $\ause$.}
\label{tab:app_eval_pix}
\end{table}
We observe improved performance for the gradient scores obtained from the predictive one-hot label with respect to both metrics. These scores are better calibrated for greater values of $p$, whereas there is no clear trend for the $\ause$ metric. In contrast, for the gradient scores obtained from the uniform label the calibration ability as well as the sparsification error enhance mostly for decreasing values of $p$.
%
\paragraph{Segment-wise Prediction Quality Estimation.}
For the segment-wise prediction quality estimation, we consider meta classification, i.e., classifying between $\IoU = 0$ (false positive) and $\IoU > 0$ (true positive), and meta regression, i.e., direct prediction of the $\IoU$. The gradient features (see eq.~\eqref{eq:meta_grads}) which are computed for each value of $p$, also separated for predictive one-hot and uniform labels, serve as input for the meta models. The results are given in Table~\ref{tab:app_eval_seg} and visualized in Figure~\ref{fig:app_metaseg}. The WideResNet backbone outperforms the SEResNeXt for meta classification and regression. Moreover, higher $\auroc$ and $R^2$ performances are achieved for greater values of $p$ independent of the architecture or the label (predictive one-hot or uniform) used for gradient scores computation. 
\begin{table}[t]
\centering
\scalebox{0.8}{
\begin{tabular}{l c cc cc}
\toprule 
\multicolumn{2}{c}{} & \multicolumn{2}{c}{last layer} & \multicolumn{2}{c}{second-to-last layer} \\
\cmidrule(r){3-4} \cmidrule(r){5-6}
 & $p$ & $\auroc$ $\uparrow$ & $R^2$ $\uparrow$ & $\auroc$ $\uparrow$ & $R^2$ $\uparrow$ \\
\midrule 
 & $0.1$             & $87.66$ & $19.40$  & $88.40$ & $29.45$ \\
Wide- & $0.3$        & $89.02$ & $38.03$  & $90.28$ & $48.23$ \\
ResNet & $0.5$       & $90.06$ & $46.82$  & $90.52$ & $49.63$ \\
(\emph{one-hot}) & $1$ & $\mathbf{91.97}$ & $\underline{50.82}$  & $91.04$ & $50.28$ \\
 & $2$                 & $\underline{91.90}$ & $50.72$  & $\mathbf{91.54}$ & $\underline{50.54}$ \\
\midrule 
 & $0.1$             & $87.97$ & $22.65$  & $89.67$ & $27.99$ \\
Wide- & $0.3$        & $89.84$ & $43.72$  & $90.84$ & $45.96$ \\
ResNet & $0.5$       & $90.66$ & $48.71$  & $91.08$ & $48.66$ \\
(\emph{uniform}) & $1$ & $91.18$ & $50.26$  & $\underline{91.36}$ & $50.48$ \\
 & $2$                 & $91.77$ & $\mathbf{51.48}$  & $\mathbf{91.54}$ & $\mathbf{51.37}$ \\
\midrule 
 & $0.1$             & $81.65$ & $12.81$  & $81.48$ & $2.29$ \\
SERes- & $0.3$       & $82.91$ & $25.73$  & $84.80$ & $32.56$ \\
NeXt & $0.5$         & $84.24$ & $33.43$  & $85.28$ & $36.21$ \\
(\emph{one-hot}) & $1$ & $85.68$ & $38.57$  & $86.01$ & $38.03$ \\
 & $2$                 & $87.70$ & $39.08$  & $87.82$ & $39.00$ \\
\midrule 
 & $0.1$             & $81.90$ & $14.41$  & $78.49$ & $6.06$ \\
SERes- & $0.3$       & $83.91$ & $30.44$  & $85.14$ & $28.28$ \\
NeXt & $0.5$         & $85.51$ & $35.40$  & $85.62$ & $33.40$ \\
(\emph{uniform}) & $1$ & $86.46$ & $37.61$  & $86.38$ & $38.05$ \\
 & $2$                 & $87.30$ & $40.54$  & $87.31$ & $40.38$ \\
\bottomrule 
\end{tabular} }
\caption{Segment-wise uncertainty evaluation results for both backbone architectures and the Cityscapes dataset as well as for the different $p$-norms and layers in terms of classification $\auroc$ and regression $R^2$.}
\label{tab:app_eval_seg}
\end{table}
%
\paragraph{OoD Segmentation.}
Our approach provides pixel-wise uncertainty scores obtained by computing the partial norm. In Figure~\ref{fig:app_heatmaps} the pixel-wise heatmaps for both backbones, different $p$-norms and the predictive one-hot as well as the uniform label are shown.
We observe that for higher $p$ values the number of uncertain pixels increases, the gradients are more sensitive to unconfident predictions. For a $p$ value of $0.1$ only a few pixels of OoD object have high uncertainty while the background is completely certain. For values of $p=1$ and $p=2$, in particular using the uniform label, the gradient scores show higher uncertainties in more sectors. To identify out-of-distribution regions, we threshold per pixel on our gradient scores, i.e., high uncertainty corresponds to out-of-distribution. Here, the OoD objects are mostly covered (and not so many background pixels) for $p$ values of $0.3$ and $0.5$. These observations are reflected in the OoD segmentation results, given in Table~\ref{tab:app_ood_obstacle} and Table~\ref{tab:app_ood_anomaly}.
\begin{table*}[t]
\centering
\resizebox{\linewidth}{!}{
\begin{tabular}{l c cc ccc cc ccc}
\toprule 
\multicolumn{2}{c}{} & \multicolumn{5}{c}{LostAndFound} & \multicolumn{5}{c}{RoadObstacle21} \\
\cmidrule(r){3-7} \cmidrule(r){8-12}
 & $p$ & AuPRC $\uparrow$ & FPR$_{95}$ $\downarrow$ & $\overline{\text{sIoU}}$ $\uparrow$ & $\overline{\text{PPV}}$ $\uparrow$ & $\overline{F_1}$ $\uparrow$ & AuPRC $\uparrow$ & FPR$_{95}$ $\downarrow$ & $\overline{\text{sIoU}}$ $\uparrow$ & $\overline{\text{PPV}}$ $\uparrow$ & $\overline{F_1}$ $\uparrow$ \\
\midrule 
          & $0.1$  & $49.5$ & $21.2$ & $39.5$ & $29.6$ & $25.1$        & $8.0$ & $23.3$ & $14.2$ & $7.5$ & $2.9$ \\
Wide-     & $0.3$ & $64.9$ & $15.9$ & $\mathbf{48.5}$ & $\underline{47.6}$ & $\underline{45.8}$         & $16.3$ & $\underline{15.4}$ & $\underline{20.8}$ & $14.4$ & $\underline{7.5}$ \\
ResNet    & $0.5$ & $64.9$ & $18.4$ & $48.3$ & $\mathbf{50.0}$ & $\mathbf{46.9}$         & $\mathbf{18.8}$ & $\mathbf{14.8}$ & $\mathbf{22.1}$ & $\underline{16.5}$ & $\mathbf{9.2}$ \\
(\emph{one-hot}) & $1$ & $44.3$ & $25.1$ & $25.8$ & $40.8$ & $21.6$    & $\underline{17.8}$ & $15.5$ & $17.1$ & $\mathbf{16.9}$ & $6.1$ \\
          & $2$   & $11.5$ & $30.6$ & $26.9$ & $26.1$ & $16.3$         & $12.8$ & $16.7$ & $19.2$ & $13.9$ & $5.5$ \\
\midrule 
          & $0.1$  & $47.1$ & $21.8$ & $38.4$ & $27.2$ & $22.3$        & $7.2$ & $25.4$ & $13.6$ & $6.9$ & $2.5$ \\
Wide-     & $0.3$ & $64.8$ & $13.5$ & $\underline{48.4}$ & $44.1$ & $43.2$         & $14.4$ & $17.2$ & $18.0$ & $13.0$ & $6.4$ \\
ResNet    & $0.5$ & $69.3$ & $9.8$ & $50.0$ & $44.8$ & $45.4$          & $16.5$ & $19.7$ & $19.5$ & $14.8$ & $7.4$ \\
(\emph{uniform}) & $1$   & $57.7$ & $10.1$ & $33.7$ & $35.2$ & $27.4$  & $8.0$ & $62.6$ & $5.7$ & $8.8$ & $1.2$ \\
          & $2$   & $8.1$ & $100.0$ & $7.5$ & $19.2$ & $4.1$           & $3.4$ & $99.9$ & $1.7$ & $14.0$ & $0.3$ \\
\midrule 
          & $0.1$  & $66.6$ & $5.2$ & $43.8$ & $35.0$ & $34.0$         & $3.5$ & $39.1$ & $5.3$ & $6.7$ & $1.1$ \\
SERes-    & $0.3$ & $\mathbf{75.1}$ & $\underline{4.2}$ & $46.2$ & $44.2$ & $43.8$          & $6.7$ & $26.8$ & $5.3$ & $9.7$ & $2.5$ \\
NeXt      & $0.5$ & $70.3$ & $7.8$ & $43.4$ & $44.6$ & $41.7$          & $8.1$ & $24.5$ & $5.8$ & $10.1$ & $3.2$ \\
(\emph{one-hot}) & $1$   & $45.1$ & $14.5$ & $22.6$ & $36.5$ & $18.5$  & $8.5$ & $24.7$ & $4.2$ & $11.8$ & $1.8$ \\
          & $2$   & $15.1$ & $18.7$ & $22.8$ & $21.9$ & $12.8$         & $5.6$ & $26.2$ & $9.0$ & $11.3$ & $3.1$ \\
\midrule 
          & $0.1$  & $64.7$ & $6.1$ & $42.8$ & $34.9$ & $32.4$         & $3.2$ & $41.3$ & $5.7$ & $6.1$ & $1.0$ \\
SERes-    & $0.3$ & $\underline{75.0}$ & $\mathbf{3.9}$ & $46.4$ & $43.8$ & $43.3$          & $6.1$ & $29.0$ & $5.2$ & $9.0$ & $2.1$ \\
NeXt      & $0.5$ & $73.8$ & $6.9$ & $45.1$ & $44.8$ & $42.3$          & $7.4$ & $28.7$ & $5.7$ & $10.3$ & $2.7$ \\
(\emph{uniform}) & $1$   & $42.0$ & $49.5$ & $17.1$ & $33.8$ & $11.7$  & $11.2$ & $79.2$ & $5.1$ & $12.0$ & $1.8$ \\
          & $2$   & $5.8$ & $100.0$ & $3.3$ & $15.7$ & $1.4$           & $9.2$ & $99.8$ & $1.7$ & $15.5$ & $1.7$ \\
\bottomrule 
\end{tabular} }
\caption{OoD segmentation results for the LostAndFound and the RoadObstacle21 dataset for different $p$-norms.}
\label{tab:app_ood_obstacle}
\end{table*}
\begin{table*}[t]
\centering
\resizebox{\linewidth}{!}{
\begin{tabular}{l c cc ccc cc ccc}
\toprule 
\multicolumn{2}{c}{} & \multicolumn{5}{c}{Fishyscapes LostAndFound} & \multicolumn{5}{c}{RoadAnomaly21} \\
\cmidrule(r){3-7} \cmidrule(r){8-12}
 & $p$ & AuPRC $\uparrow$ & FPR$_{95}$ $\downarrow$ & $\overline{\text{sIoU}}$ $\uparrow$ & $\overline{\text{PPV}}$ $\uparrow$ & $\overline{F_1}$ $\uparrow$ & AuPRC $\uparrow$ & FPR$_{95}$ $\downarrow$ & $\overline{\text{sIoU}}$ $\uparrow$ & $\overline{\text{PPV}}$ $\uparrow$ & $\overline{F_1}$ $\uparrow$ \\
\midrule 
          & $0.1$  & $23.9$ & $22.0$ & $13.9$ & $29.5$ & $15.6$      & $28.2$ & $75.4$ & $18.2$ & $14.8$ & $4.2$ \\
Wide-     & $0.3$ & $\underline{26.9}$ & $31.2$ & $\mathbf{16.2}$ & $30.0$ & $\underline{18.1}$       & $33.8$ & $70.5$ & $20.9$ & $18.4$ & $6.0$ \\
ResNet    & $0.5$ & $22.8$ & $35.5$ & $12.1$ & $27.3$ & $14.1$       & $34.5$ & $69.5$ & $19.4$ & $19.3$ & $5.5$ \\
(\emph{one-hot}) & $1$   & $10.1$ & $39.1$ & $4.0$ & $14.6$ & $3.3$  & $32.5$ & $69.5$ & $16.1$ & $17.0$ & $5.8$ \\
          & $2$   & $1.5$ & $41.7$ & $11.5$ & $2.7$ & $1.7$          & $25.7$ & $70.2$ & $15.5$ & $15.1$ & $5.8$ \\
\midrule 
          & $0.1$  & $23.0$ & $\underline{21.7}$ & $13.6$ & $27.6$ & $14.7$     & $27.6$ & $75.9$ & $17.8$ & $14.6$ & $4.1$ \\
Wide-     & $0.3$ & $\mathbf{28.3}$ & $29.2$ & $\underline{15.7}$ & $\underline{32.1}$ & $\mathbf{18.4}$      & $33.7$ & $69.7$ & $20.8$ & $17.1$ & $5.8$ \\
ResNet    & $0.5$ & $\underline{26.9}$ & $36.6$ & $14.8$ & $29.6$ & $16.5$      & $36.7$ & $61.4$ & $21.6$ & $17.8$ & $6.2$ \\
(\emph{uniform}) & $1$   & $0.8$ & $66.4$ & $6.2$ & $2.5$ & $2.0$   & $\underline{45.2}$ & $\underline{60.7}$ & $24.8$ & $26.2$ & $\underline{9.5}$ \\
          & $2$   & $0.2$ & $99.8$ & $0.2$ & $0.3$ & $0.0$          & $29.2$ & $97.7$ & $21.1$ & $\underline{29.0}$ & $6.1$ \\
\midrule 
          & $0.1$  & $21.0$ & $\mathbf{15.6}$ & $10.4$ & $21.2$ & $8.3$      & $36.2$ & $65.7$ & $21.9$ & $16.9$ & $5.7$ \\
SERes-    & $0.3$ & $23.7$ & $23.7$ & $10.1$ & $26.8$ & $11.6$      & $40.5$ & $64.9$ & $24.5$ & $19.5$ & $8.7$ \\
NeXt      & $0.5$ & $20.8$ & $29.8$ & $8.1$ & $23.9$ & $9.4$        & $39.3$ & $66.5$ & $23.1$ & $21.5$ & $7.8$ \\
(\emph{one-hot}) & $1$   & $8.8$ & $36.3$ & $4.6$ & $12.4$ & $3.4$  & $33.3$ & $69.4$ & $17.2$ & $16.2$ & $8.1$ \\
          & $2$   & $1.2$ & $39.4$ & $11.5$ & $2.2$ & $1.6$         & $27.1$ & $71.0$ & $16.2$ & $14.0$ & $7.4$ \\
\midrule 
          & $0.1$  & $20.6$ & $\mathbf{15.6}$ & $8.1$ & $21.6$ & $6.6$       & $35.6$ & $66.2$ & $21.3$ & $17.0$ & $5.3$ \\
SERes-    & $0.3$ & $24.1$ & $21.8$ & $10.5$ & $26.3$ & $11.9$      & $41.1$ & $62.7$ & $\underline{24.9}$ & $20.0$ & $8.5$ \\
NeXt      & $0.5$ & $22.6$ & $32.0$ & $8.6$ & $\mathbf{32.3}$ & $11.0$       & $42.8$ & $\mathbf{56.4}$ & $\mathbf{25.8}$ & $21.8$ & $\mathbf{9.7}$ \\
(\emph{uniform}) & $1$   & $0.6$ & $83.3$ & $1.8$ & $2.3$ & $0.6$   & $\mathbf{47.4}$ & $67.3$ & $23.7$ & $24.9$ & $9.0$ \\
          & $2$   & $0.2$ & $99.8$ & $0.2$ & $0.3$ & $0.0$          & $35.0$ & $98.1$ & $22.3$ & $\mathbf{30.7}$ & $8.3$ \\
\bottomrule 
\end{tabular} }
\caption{OoD segmentation results for the Fishyscapes LostAndFound and the RoadAnomaly21 dataset for different $p$-norms.}
\label{tab:app_ood_anomaly}
\end{table*}
For the LostAndFound dataset and the Fishyscapes LostAndFound dataset, the best results are achieved for the $0.3$ and $0.5$ $p$-norms. For the RoadAnomaly21 dataset, also for higher $p$ values strong (in one case even the best) results are obtained. Across these three datasets, there is no favorability which backbone architecture or label (predictive one-hot or uniform) performs better. In comparison for the RoadObstacle21 dataset, the WideResNet backbone with gradient scores obtained from the predictive one-hot performs best.

In summary, there is no clear tendency which $p$-norm outperforms the others for the different tasks of pixel- and segment-wise uncertainty estimation as well as for OoD segmentation. 
However, there is a strong trend towards $p \in \{0.3, 0.5\}$ performing especially strongly.
%
%
\section{Ablation: Deeper Layers}\label{sec:app_E}
In Section~\ref{sec:deep_grad}, we have shown how to compute the gradients also for deeper layers than the last convolutional one. In this section, we discuss the numerical results for the second-to-last layer in comparison to the last.
%
\paragraph{Pixel-wise Uncertainty Evaluation.}
The results for the pixel-wise uncertainty estimation measured by $\ece$ and $\ause$ are given in Table~\ref{tab:app_eval_pix} for both layers. For the gradient scores obtained from the predictive one-hot, the evaluation metrics are mainly similar showing only small performance gaps. The results differ for the uniform labels, although there is no trend to which layer achieves the higher ones. 
%
\paragraph{Segment-wise Prediction Quality Estimation.}
The segment-wise meta classification and regression results are shown in Table~\ref{tab:app_eval_seg}. We observe the same behavior for the second-to-last as for the last one, namely that as $p$ increases, performance improves for both metrics. Furthermore, the performance is almost equal for the second-to-last and the last layer for the $1$ and the $2$ norm independent of the backbone and labels to obtain the gradient scores.
%
\paragraph{OoD Segmentation.}
In Table~\ref{tab:app_ood_stl} the OoD segmentation results are given. 
\begin{table*}[t]
\centering
\resizebox{\linewidth}{!}{
\begin{tabular}{l c cc ccc cc ccc}
\toprule 
\multicolumn{2}{c}{} & \multicolumn{5}{c}{LostAndFound} & \multicolumn{5}{c}{Fishyscapes LostAndFound} \\
\cmidrule(r){3-7} \cmidrule(r){8-12}
 & $p$ & AuPRC $\uparrow$ & FPR$_{95}$ $\downarrow$ & $\overline{\text{sIoU}}$ $\uparrow$ & $\overline{\text{PPV}}$ $\uparrow$ & $\overline{F_1}$ $\uparrow$ & AuPRC $\uparrow$ & FPR$_{95}$ $\downarrow$ & $\overline{\text{sIoU}}$ $\uparrow$ & $\overline{\text{PPV}}$ $\uparrow$ & $\overline{F_1}$ $\uparrow$ \\
\midrule 
          & $0.1$  & $10.3$ & $40.0$ & $12.6$ & $15.8$ & $4.0$         & $\mathbf{2.0}$ & $\underline{36.8}$ & $1.6$ & $3.0$ & $0.6$ \\
Wide-     & $0.3$ & $\mathbf{10.9}$ & $32.9$ & $23.1$ & $23.0$ & $11.3$         & $1.7$ & $39.4$ & $2.3$ & $2.2$ & $0.5$ \\
ResNet    & $0.5$ & $\underline{10.5}$ & $32.4$ & $26.9$ & $25.5$ & $15.3$         & $1.5$ & $40.3$ & $9.6$ & $2.3$ & $1.3$ \\
(\emph{one-hot}) & $1$   & $10.0$ & $32.2$ & $\underline{28.9}$ & $\mathbf{26.1}$ & $\underline{17.3}$  & $1.4$ & $41.1$ & $\underline{12.2}$ & $2.5$ & $\underline{1.9}$ \\
          & $2$   & $9.7$ & $32.1$ & $\mathbf{30.0}$ & $\underline{25.7}$ & $\mathbf{18.3}$    & $1.3$ & $41.5$ & $\mathbf{13.6}$ & $2.8$ & $\mathbf{2.4}$ \\
\midrule 
          & $0.1$  & $0.7$ & $98.8$ & $0.5$ & $0.8$ & $0.0$            & $0.3$ & $96.0$ & $0.8$ & $0.7$ & $0.0$ \\
Wide-     & $0.3$ & $0.5$ & $99.8$ & $0.5$ & $0.8$ & $0.0$             & $0.2$ & $98.9$ & $0.2$ & $0.3$ & $0.0$ \\
ResNet    & $0.5$ & $0.5$ & $99.9$ & $0.5$ & $0.8$ & $0.0$             & $0.2$ & $99.5$ & $0.2$ & $0.3$ & $0.0$ \\
(\emph{uniform}) & $1$   & $0.5$ & $100.0$ & $0.5$ & $0.8$ & $0.0$     & $0.2$ & $99.7$ & $0.2$ & $0.3$ & $0.0$ \\
          & $2$   & $0.5$ & $100.0$ & $0.5$ & $0.8$ & $0.0$            & $0.2$ & $99.8$ & $0.2$ & $0.3$ & $0.0$ \\
\midrule 
          & $0.1$  & $3.0$ & $66.2$ & $4.9$ & $7.3$ & $1.4$            & $\underline{1.9}$ & $41.1$ & $5.8$ & $\mathbf{5.2}$ & $1.8$ \\
SERes-    & $0.3$ & $6.8$ & $27.1$ & $15.4$ & $12.9$ & $6.1$           & $\underline{1.9}$ & $\mathbf{36.0}$ & $5.2$ & $\underline{5.0}$ & $1.6$ \\
NeXt      & $0.5$ & $8.1$ & $23.9$ & $19.5$ & $15.2$ & $8.6$           & $1.7$ & $\underline{36.8}$ & $4.1$ & $4.3$ & $1.0$ \\
(\emph{one-hot}) & $1$   & $9.3$ & $\underline{21.7}$ & $22.6$ & $17.8$ & $11.1$   & $1.4$ & $37.6$ & $7.0$ & $2.1$ & $0.9$ \\
          & $2$   & $10.4$ & $\mathbf{20.4}$ & $24.1$ & $18.4$ & $12.7$         & $1.2$ & $38.3$ & $11.5$ & $2.3$ & $1.7$ \\
\midrule 
          & $0.1$  & $0.4$ & $99.9$ & $0.5$ & $0.8$ & $0.0$            & $0.2$ & $88.9$ & $0.3$ & $0.0$ & $0.0$ \\
SERes-    & $0.3$ & $0.4$ & $99.9$ & $0.5$ & $0.8$ & $0.0$             & $0.2$ & $93.2$ & $0.2$ & $0.0$ & $0.0$ \\
NeXt      & $0.5$ & $0.4$ & $100.0$ & $0.5$ & $0.8$ & $0.0$            & $0.2$ & $96.1$ & $0.2$ & $0.3$ & $0.0$ \\
(\emph{uniform}) & $1$   & $0.4$ & $100.0$ & $0.5$ & $0.8$ & $0.0$     & $0.2$ & $99.2$ & $0.2$ & $0.3$ & $0.0$ \\
          & $2$   & $0.5$ & $100.0$ & $0.5$ & $0.8$ & $0.0$            & $0.2$ & $99.7$ & $0.2$ & $0.3$ & $0.0$ \\
\bottomrule 
\end{tabular} }
\caption{OoD segmentation results for the LostAndFound and the Fishyscapes LostAndFound dataset for different $p$-norms and the second-to-last layer.}
\label{tab:app_ood_stl}
\end{table*}
In comparison to the performance of the gradient scores of the last layer (see Table~\ref{tab:app_ood_obstacle} and Table~\ref{tab:app_ood_anomaly}), the performance of the second-to-last layer is poor, i.e., the results for all evaluation metrics are worse than these of the last layer. In some cases, there is no detection capability at all for the gradient scores obtained from the uniform label.

In conclusion, the gradients of the second-to-last layer do not improve the uncertainty estimation (at pixel- and segment-level) nor the OoD segmentation quality, rather they perform worse in some cases.
The finding that deeper layer gradients contain less information than the final layer has been observed before outside the semantic segmentation setting in \cite{Huang2021} and \cite{Riedlinger2023}.

\end{document}